\pdfoutput=1

\documentclass[11pt]{article}

\usepackage[final]{acl}

\usepackage{times}
\usepackage{latexsym}

\usepackage[T1]{fontenc}

\usepackage[utf8]{inputenc}

\usepackage{microtype}

\usepackage{inconsolata}

\usepackage{graphicx}

\usepackage{makecell}

\usepackage{amsfonts}

\usepackage{booktabs}

\usepackage{diagbox}

\usepackage{colortbl}

\usepackage{xcolor}

\usepackage{array}

\usepackage{multirow}

\usepackage[ruled]{algorithm2e}

\usepackage{epsfig}

\usepackage{amsmath}
\usepackage{arydshln}
\usepackage{nameref}

\title{FISH-Tuning: Enhancing PEFT Methods with Fisher Information}

\author{
 \textbf{Kang Xue\textsuperscript{1,2,3}},
 \textbf{Ming Dong\textsuperscript{1,2,3}},
 \textbf{Xinhui Tu\textsuperscript{1,2,3}},
 \textbf{Tingting He\textsuperscript{1,2,3}*},
 \\
 \textsuperscript{1}Hubei Provincial Key Laboratory of Artificial Intelligence and Smart Learning
 \\
 \textsuperscript{2}National Language Resources Monitoring and Research Center for Network Media
 \\
 \textsuperscript{3}School of Computer, Central China Normal University, Wuhan, China
 \\
 \textsuperscript{*}Corresponding author
 \\
 \href{mailto:xuekang@mails.ccnu.edu.cn}{xuekang@mails.ccnu.edu.cn}
 \\
 \href{mailto:dongming@ccnu.edu.cn,tuxinhui@ccnu.edu.cn,tthe@ccnu.edu.cn}{\{dongming,tuxinhui,tthe\}@ccnu.edu.cn}
}

\begin{document}
\maketitle
\begin{abstract}

The rapid growth in the parameter size of Large Language Models (LLMs) has spurred the development of Parameter-Efficient Fine-Tuning (PEFT) 
methods to mitigate the substantial computational costs of fine-tuning. Among these, Fisher Induced Sparse uncHanging (FISH) Mask is a selection-based 
PEFT technique that identifies a critical subset of pre-trained parameters using approximate Fisher information. 
While addition-based and reparameterization-based PEFT methods like LoRA and Adapter already fine-tune only a small number of parameters, 
the newly introduced parameters within these methods themselves present an opportunity for further optimization. 
Selectively fine-tuning only the most impactful among these new parameters could further reduce resource consumption while maintaining, 
or even improving, fine-tuning effectiveness. In this paper, we propose \textbf{FISH-Tuning}, a novel approach that incorporates FISH Mask 
into such PEFT methods, including LoRA, Adapter, and their variants. By leveraging Fisher information to identify and update only 
the most significant parameters within these added or reparameterized components, FISH-Tuning aims to achieve superior performance without 
increasing training time or inference latency compared to the vanilla PEFT methods. Experimental results across 
various datasets and pre-trained models demonstrate that FISH-Tuning consistently outperforms the vanilla PEFT methods when using the 
same proportion of trainable parameters. Code is available at \href{https://anonymous.4open.science/r/FISH-Tuning-6F7C}{https://anonymous.4open.science/r/FISH-Tuning-6F7C}.

\end{abstract}

\section{Introduction}


The emergence of Large Language Models (LLMs) has revolutionized Natural Language Processing (NLP) by achieving remarkable performance across a wide range of tasks. These models, typically trained on massive datasets using self-supervised learning, are fine-tuned on downstream tasks through processes such as Supervised Fine-Tuning (SFT) and Reinforcement Learning with Human Feedback (RLHF) \cite{DBLP:conf/nips/ChristianoLBMLA17, DBLP:journals/corr/abs-2009-01325, DBLP:conf/nips/Ouyang0JAWMZASR22}. However, fine-tuning all parameters of such large LLMs is computationally expensive, requiring substantial GPU memory and training time. This challenge has led to the rise of Parameter-Efficient Fine-Tuning (PEFT) \cite{DBLP:journals/natmi/DingQYWYSHCCCYZWLZCLTLS23} methods, which aim to achieve competitive performance by fine-tuning only a small subset of parameters.


PEFT methods can be broadly categorized into three types \cite{DBLP:journals/corr/abs-2303-15647,DBLP:journals/corr/abs-2403-14608}: (1) \textbf{Selection-based methods}, which fine-tune a subset of pre-trained parameters while freezing the rest, e.g., BitFit \cite{DBLP:conf/acl/ZakenGR22}, Diff-Pruning \cite{DBLP:conf/acl/GuoRK20}, and FISH Mask \cite{DBLP:conf/nips/SungNR21}. (2) \textbf{Addition-based methods}, which introduce additional trainable parameters or layers into the model, e.g., Adapters \cite{DBLP:conf/icml/HoulsbyGJMLGAG19}, $(\text{IA})^3$ \cite{DBLP:conf/nips/LiuTMMHBR22}, and Prefix-Tuning \cite{DBLP:conf/acl/LiL20}. (3) \textbf{Reparameterization-based methods}, which use low-rank representations to reduce the number of trainable parameters, e.g., LoRA \cite{DBLP:conf/iclr/HuSWALWWC22}, DoRA \cite{DBLP:conf/icml/LiuWY0WCC24}, and IntrinsicSAID \cite{DBLP:conf/acl/AghajanyanGZ20}. While these methods have proven effective individually, hybrid approaches that combine multiple PEFT techniques are gaining attention to further improve efficiency and performance.



FISH Mask is a parameter-efficient fine-tuning (PEFT) method that selects important pre-trained model parameters for fine-tuning using Fisher information. 
In contrast, common PEFT methods such as LoRA typically update all parameters within their added modules, like the low-rank matrices. 
This approach does not consider that these newly introduced parameters may also vary in importance for the downstream task. 
We propose that Fisher information can guide the selection of which PEFT parameters to update. This concept has received little attention so far. 
By applying selective fine-tuning to the internal structure of PEFT modules, 
it may be possible to improve both the effectiveness and efficiency of these techniques.


In this paper, we propose \textbf{FISH-Tuning}, a novel framework that integrates FISH Mask into addition-based and reparameterization-based PEFT methods, including LoRA, Adapters, and their variants. FISH-Tuning leverages Fisher information to select the most critical parameters within these methods, enabling efficient fine-tuning without additional memory overhead or inference latency. Specifically, we demonstrate how FISH Mask can be applied to LoRA, DoRA, Adapters, Prefix-Tuning, and $(\text{IA})^3$. Experimental results across multiple datasets and pre-trained models show that FISH-Tuning consistently outperforms the original PEFT methods with the same proportion of trainable parameters. We summarize our contributions as follows:

\begin{itemize}
\item We introduce FISH-Tuning, a novel framework that integrates FISH Mask into addition-based and reparameterization-based PEFT methods, enabling efficient parameter selection without increasing memory or latency.
\item We demonstrate the effectiveness of FISH-Tuning across various datasets and pre-trained models, achieving consistent performance improvements over the original PEFT methods.
\item We provide insights into the role of Fisher information in parameter selection, offering a new perspective on optimizing PEFT methods.
\end{itemize}


    




\section{Related Work}

\subsection{Sparsely Updating Parameter-Efficient Fine-tuning}
Parameter-Efficient Fine-Tuning (PEFT) methods aim to adapt large pre-trained language models with minimal parameter updates, 
and a subset of these approaches leverages sparsity to further enhance efficiency. Sparse updating in PEFT focuses on selectively modifying 
a small fraction of parameters or enforcing sparse structures during adaptation, reducing computational overhead and storage requirements 
while maintaining performance.

One prominent method in this domain is LoRI \cite{zhang2025lori}. Unlike conventional LoRA, which learns dense 
low-rank matrices $ A $ and $ B $, LoRI freezes $ A $ as a random projection and sparsely updates only the most task-relevant entries in $ B $. 
This design eliminates the need to store gradients and optimizer states for $ A $, significantly reducing memory consumption while preserving 
performance through task-specific masking. Moreover, LoRI introduces an additional layer of efficiency by applying global magnitude-based sparsity to $ B $, 
allowing up to 90\% of its parameters to remain untouched during adaptation. By focusing updates on a small, 
critical subset of parameters, LoRI achieves strong performance while drastically lowering trainable parameter counts—up to 95\% fewer than standard LoRA in some settings.


\subsection{Fisher Information}
\label{sec:fisher}
The Fisher Information Matrix (FIM) \cite{fisher1922mathematical,amari1996neural} is a fundamental tool in deep learning neural networks that measures parameter importance and helps address catastrophic forgetting \cite{french1999catastrophic,mccloskey1989catastrophic,mcclelland1995there,ratcliff1990connectionist}. Its three key advantages are \cite{DBLP:journals/corr/abs-1301-3584}: approximates the Hessian matrix near loss function minima; can be computed efficiently using first-order derivatives; is positive semi-definite, ensuring stable optimization. These properties make it particularly valuable for techniques like Elastic Weight Consolidation (EWC), which uses FIM to identify and protect important parameters while learning new tasks, thus helping preserve previously learned information.

\section{Preliminary}




\subsection{Task Definition}
The goal of PEFT is to fine-tune the model with as few trainable parameters as possible while achieving better results. Therefore, we can compare our method with the original PEFT method using the same trainable parameter ratio and evaluate their performance on the same dataset and hyperparameters.

\subsection{FISH Mask Based PEFT}

Fisher Information Matrix (FIM) is defined as:
\begin{equation}
    \resizebox{0.85\linewidth}{!}{$
    F_\theta = 	\mathbb{E}_{x \sim p(x)} [ \mathbb{E}_{y \sim p_\theta(y|x)} \nabla_\theta \log p_\theta(y|x) \nabla_\theta \log p_\theta(y|x)^\textrm{T} ]
    $,}
\end{equation}

where $x$ is the input, $y$ is the output, $\theta$ represents the model's parameters, $p(x)$ is the probability distribution of the input $x$, and $\nabla$ is the gradient.
Fisher information is typically estimated using a diagonal approximation, in which gradients for all parameters are calculated based on $N$ data samples:
\begin{align}
\hat{F}_{\theta} &= \frac{1}{N} \sum_{i=1}^{N} \mathbb{E}_{y \sim p_{\theta}(y|x_i)} \Big[ \nabla_{\theta} \log p_{\theta}(y|x_i) \nonumber \\
&\qquad \odot \nabla_{\theta} \log p_{\theta}(y|x_i) \Big],
\end{align}
where $N$ is the number of data samples, and $\odot$ is the Hadamard product. In supervised learning, we can use ``Empirical Fisher information'' for further approximation:
\begin{equation}
    \resizebox{0.85\linewidth}{!}{$
    \hat{F}_{\theta} = \frac{1}{N} \sum_{i=1}^{N}  \nabla_{\theta} \log p_{\theta}(y_i|x_i) \odot \nabla_{\theta} \log p_{\theta}(y_i|x_i)
    $}
    \label{eq:empfisher}
\end{equation}
We will select the top-k parameters $\theta_i$ according to the estimated FIM:
\begin{equation}
\label{eq:thetaselect}
\theta_{\text{selected}} = \{ \theta_i \ | \ \hat{F}_{\theta_i} \ge \text{sort}(\hat{F}_{\theta})_k \}
\end{equation}
Then we create the binary mask $M$ based on the top-k importance values in $\hat{F}_{\theta}$:
\begin{equation}
M_{i} =
\begin{cases}
1, & \text{if } \theta_i \in \theta_{\text{selected}} \\
0, & \text{otherwise}
\end{cases}
\end{equation}
Finally, we mask the gradients for the loss function:
\begin{equation}
\nabla_{\theta_i} \mathcal{L}^{\text{masked}} = (\nabla_{\theta_i} \mathcal{L} )\odot M_i
\end{equation}
The masked gradients can be used to update the parameters $\theta_i$ using Stochastic Gradient Descent (SGD) or Adam \cite{DBLP:journals/corr/KingmaB14} optimizer.

\section{Method}
\label{sec:method}

\begin{figure}[htbp]
    \centering
    \begin{minipage}[c]{1\linewidth}
        \centering
        \includegraphics[width=1.0\linewidth]{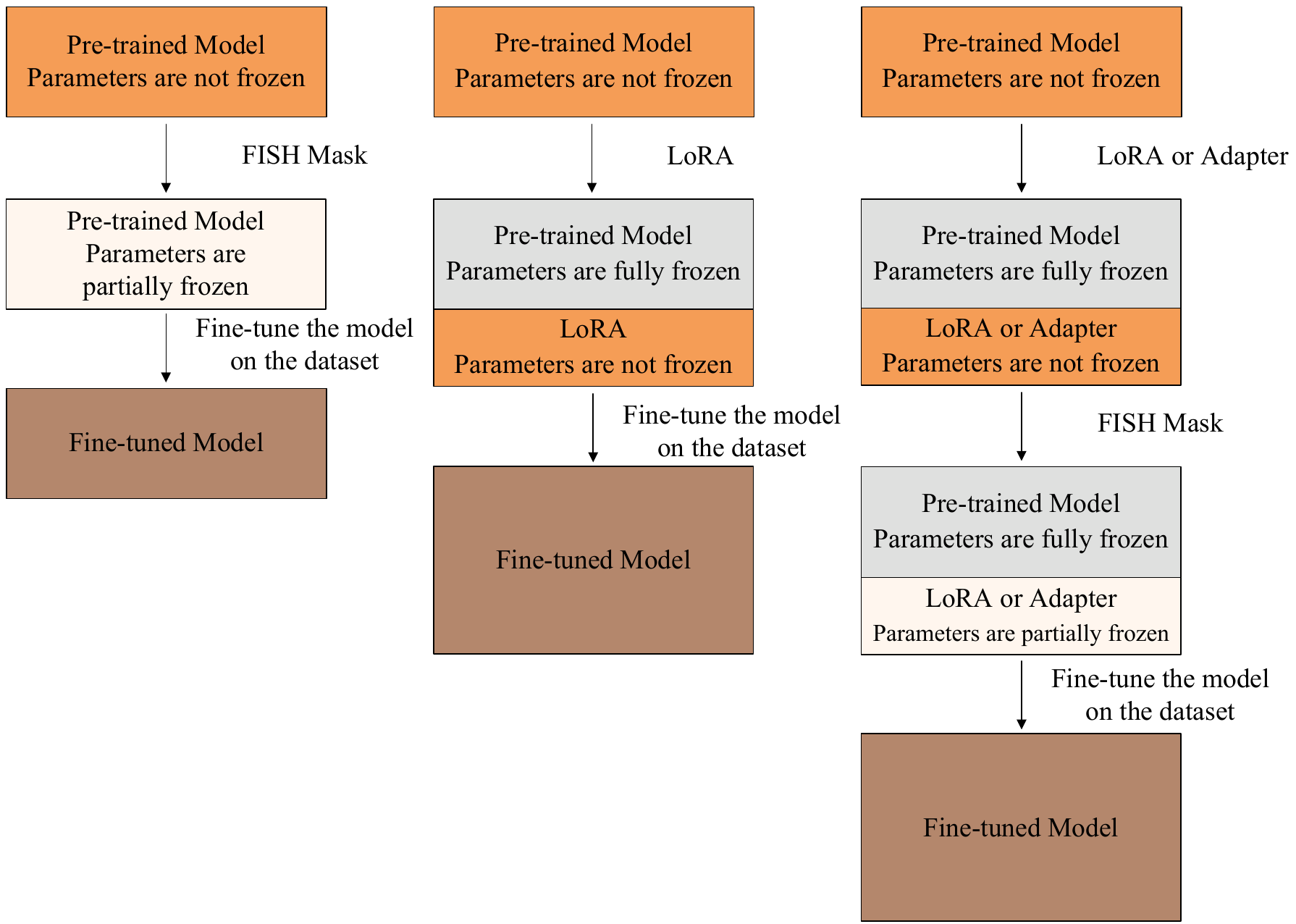}
    \end{minipage} 
    \caption{Original FISH Mask method (left) and original LoRA method (mid) and our FISH-Tuning method (right).}
    \label{fig:fishtuningintro}
\end{figure}

We use FISH Mask into the Addition-based methods like Adapter, Prefix-Tuning, $(\text{IA})^3$, and the Reparameterization-based methods like LoRA, DoRA. 
We also use FISH Mask into the Hybrid PEFT method like UniPELT \cite{DBLP:conf/acl/MaoMHAM0YK22}. 
The differences between our method and the original FISH Mask and LoRA are illustrated in Fig.~\ref{fig:fishtuningintro}.
We believe that our method can also be used in other PEFT methods.

\subsection{FISH Mask in Reparameterization-based Methods}
\subsubsection{FISH Mask in LoRA}
\label{sec:lora}

\begin{figure}[htbp]
    \centering
    \begin{minipage}[c]{1\linewidth}
        \centering
        \includegraphics[width=1.0\linewidth]{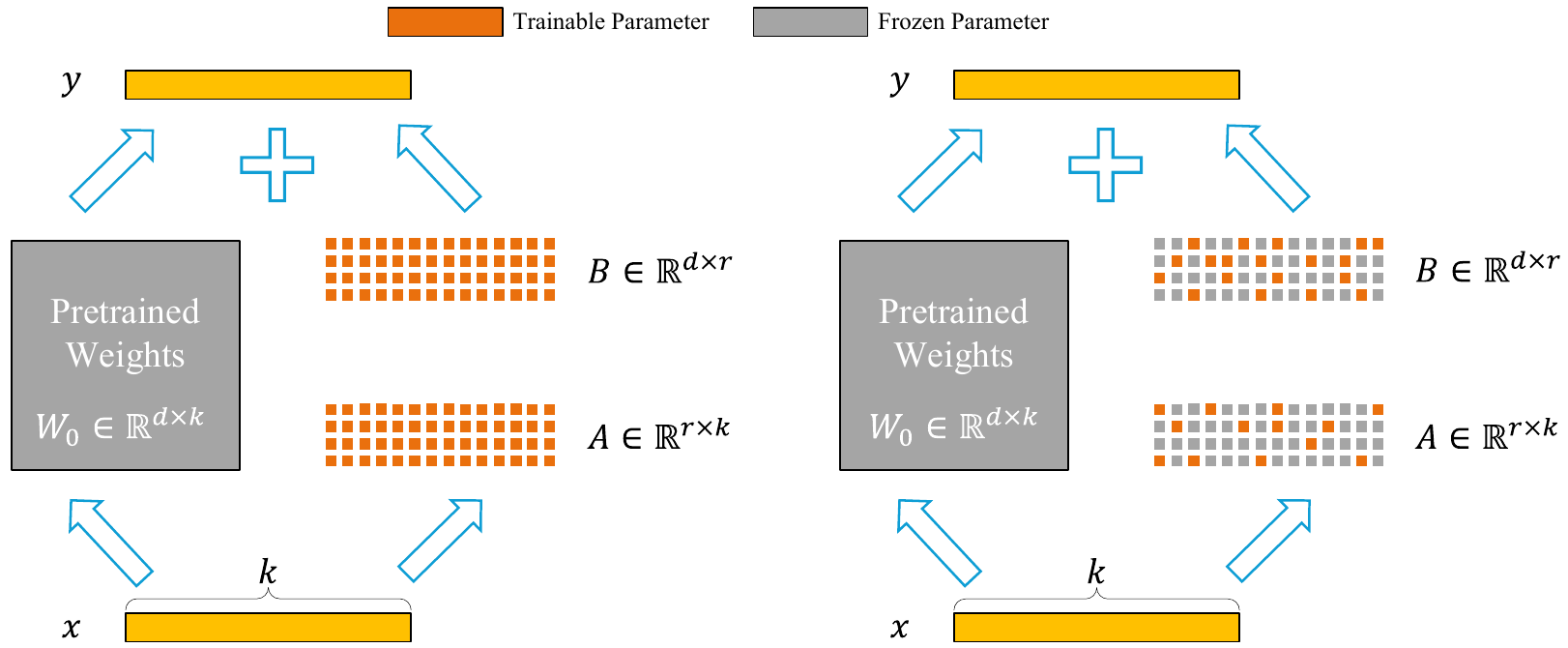}
    \end{minipage} 
    \caption{Original LoRA method (left) and the LoRA-FISH method (right).}
    \label{fig:lora}
\end{figure}

In the original LoRA method, the update to the weight matrix  $W_0 \in \mathbb{R}^{d \times k}$ is represented as:
\begin{equation}
  \label{eq:loraorigin}
  W_0 + \Delta W = W_0 + BA,
\end{equation}
where  $B \in \mathbb{R}^{d \times r}$  and  A $\in \mathbb{R}^{r \times k}$ , and the update involves training the matrices  $B$  and  $A$  while  $W_0$  is frozen.  

For the FISH-Tuning method in LoRA, we define the combined vector of  B  and  A  as $ \tilde{\theta} \in \mathbb{R}^{d \times r + r \times k} $. Then we use Eq.~\ref{eq:empfisher} to calculate the importance score of $\tilde{\theta}$ and create the related binary mask for it. 

The difference between the original LoRA method and the FISH-Tuning method is shown in Fig.~\ref{fig:lora}.

\subsubsection{FISH Mask in DoRA}

In the original DoRA method, the update to the weight matrix $W_0 \in \mathbb{R}^{d \times k}$  is represented as:
\begin{equation}
  \label{eq:doraorigin}
  W_0 + \Delta W = m \frac{W_0 + BA}{\left\| W_0 + BA \right\|_c},
\end{equation}
where $m \in \mathbb{R}^{1 \times k}$,  $B \in \mathbb{R}^{d \times r}$  ,  $A \in \mathbb{R}^{r \times k}$, $\left\| . \right\|_c$ is the vector-wise norm of a matrix across each column , and the update involves training the matrices  $B$, $A$ and vector $m$  while  $W_0$  is frozen.  

For the FISH-Tuning method in DoRA, we define the combined vector of  $B$, $A$, and $m$  as $ \tilde{\theta} \in \mathbb{R}^{1 \times k + d \times r + r \times k} $. Then we use Eq.~\ref{eq:empfisher} to calculate the importance score of $\tilde{\theta}$ and create the related binary mask for it.

\subsection{FISH Mask in Addition-based Methods}
\subsubsection{FISH Mask in Adapter}
\label{sec:adapter}

\begin{figure}[htbp]
    \centering
    \begin{minipage}[c]{1\linewidth}
        \centering
        \includegraphics[width=1.0\linewidth]{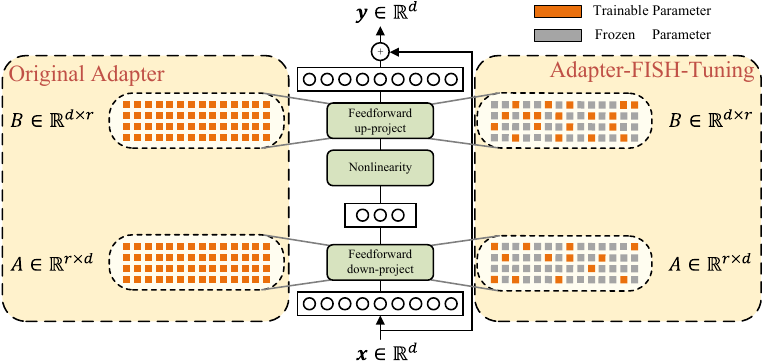}
    \end{minipage} 
    \caption{Original Adapter method (left) and the Adapter-FISH method (right).}
    \label{fig:adapter}
\end{figure}

In the original Serial Adapter method, it adds the adapter module twice to each Transformer layer: after the projection following multi-head attention and after the two feed-forward layers. The formula can be represented as:
\begin{equation}
  \label{eq:adapterorigin}
  \text{Adapter}(\boldsymbol{x}) = B \sigma(A\boldsymbol{x}) + \boldsymbol{x},
\end{equation}
where $B \in \mathbb{R}^{d \times r}$, $A \in \mathbb{R}^{r \times d}$, $\boldsymbol{x} \in \mathbb{R}^{d}$, $\sigma$ is the non-linear activation function. The update involves training the matrices $B$ and $A$ while $x$ is frozen.

For the FISH-Tuning method in Adapter, we define the combined vector of  $B$ and $A$  as $ \tilde{\theta} \in \mathbb{R}^{ d \times r + r \times d} $. Then we use Eq.~\ref{eq:empfisher} to calculate the importance score of $\tilde{\theta}$ and create the related binary mask for it.

The difference between the original Adapter method and the FISH-Tuning method is shown in Fig.~\ref{fig:adapter}.

\subsubsection{FISH Mask in Prefix-Tuning}
\label{sec:prefixtuning}

Prefix Tuning introduces new parameters into the multi-head attention blocks in each Transformer layer. More specifically, it prepends trainable prefix vectors $P^K$ and $P^V$ to the keys and values of the attention head input, each with a configurable prefix length $l$:
\begin{equation}
  \resizebox{0.85\linewidth}{!}{$
  \text{head}_i=\text{Attention}(QW_i^Q, [P_i^K,KW_i^K], [P_i^V,VW_i^V])
  $,}
  \label{eq:prefixorigin}
\end{equation}
where $W_i \in \mathbb{R}^{d \times k } $, $Q, K, V \in \mathbb{R}^{seq\_len \times d }$, $P_i \in \mathbb{R}^{l \times k }$, and $[P_i^K, KW_i^K]\in \mathbb{R}^{(l+seq\_len) \times k}$. $head_i$ is the $i$-th attention head. The update involves training the matrices $P_i$ while $W_i$ is frozen.

For the FISH-Tuning method in Prefix-Tuning, we define the combined vector of  $P_i^K$ and $P_i^V$  as $ \tilde{\theta} \in \mathbb{R}^{ 2 \times l \times k} $. Then we use Eq.~\ref{eq:empfisher} to calculate the importance score of $\tilde{\theta}$ and create the related binary mask for it.

\subsubsection{FISH Mask in $(\text{IA})^3$}
In the original $(\text{IA})^3$ method, it introduces trainable vectors  into different components of a Transformer model, which perform element-wise rescaling of inner model activations. The formula can be represented as:
\begin{equation}
    \resizebox{0.85\linewidth}{!}{$
    \text{head}_i=\text{Attention}(QW_i^Q, l_k \odot  KW_i^K, l_v \odot VW_i^V)
    \label{eq:ia3origin}
    $}
\end{equation}
\begin{equation}
  \label{eq:ia3origin2}
  \text{FFN}(x)=(l_{ff} \odot \sigma (Ax))B,
\end{equation}
where $W_i \in \mathbb{R}^{d \times k } $, $Q, K, V \in \mathbb{R}^{seq\_len \times d }$, $B \in \mathbb{R}^{d \times r}$, $A \in \mathbb{R}^{r \times d}$, $x \in \mathbb{R}^{d \times k}$,  $l \in \mathbb{R}^{k }$, and $ l_k \odot  KW_i^K \in \mathbb{R}^{seq\_len \times k}$. $head_i$ is the $i$-th attention head. $FFN$ is the Feed-forward Network. $\sigma$ is the non-linear activation function. The update involves training the vectors $l$ while $W_i$, $A$, and $B$ are frozen.

For the FISH-Tuning method in $(\text{IA})^3$, we define the combined vector of  $l_k$, $l_v$ and $l_{ff}$  as $ \tilde{\theta} \in \mathbb{R}^{ 3 \times k} $. Then we use Eq.~\ref{eq:empfisher} to calculate the importance score of $\tilde{\theta}$ and create the related binary mask for it.

\subsection{FISH Mask in UniPELT}

In the original UniPELT method, it adds a trainable gating value ${\mathcal{G}_m} \in (0, 1)$ that is computed via a feed-forward network $W_{\mathcal{G}_m}$ and sigmoid activation $\sigma$ from the Transformer layer input states $x$:
\begin{equation}
  \label{eq:unipelt}
  \mathcal{G}_m=\sigma(W_{\mathcal{G}_m}x)
\end{equation}
These gating values are then used to scale the output activations of the injected PEFT modules, e.g., for a LoRA layer:
\begin{equation}
  \label{eq:unipelt2}
  W_0+\Delta W=W_0+ \mathcal{G}_{LoRA} BA,
\end{equation}
where the update involves training the matrices $W_{\mathcal{G}_m}$,  $B$,  and  $A$  while  $W_0$  is frozen.  


In our settings for UniPELT, we use LoRA, Adapter, Prefix-Tuning as modules and add separate gating values for them. For the FISH-Tuning method, we follow the settings from Character \ref{sec:lora} for the LoRA component, Character \ref{sec:adapter} for the Adapter component, and Character \ref{sec:prefixtuning} for the Prefix-Tuning component.

\section{Experiments Setup}

\subsection{Datasets and Baselines}
\textbf{Datasets.}
\label{datasets}
We conduct experiments on two different types of tasks: classification tasks and generation tasks. For classification tasks, we select the GLUE \cite{DBLP:conf/iclr/WangSMHLB19} dataset, 
and for generation tasks, we choose the GSM8K \cite{DBLP:journals/corr/abs-2110-14168} dataset. For the GLUE dataset, we use BERT \cite{DBLP:conf/naacl/DevlinCLT19}, 
Modern BERT \cite{DBLP:journals/corr/abs-2412-13663}, and LLaMA-3.2-1B \cite{DBLP:journals/corr/abs-2407-21783} as the pre-trained models. For the GSM8K dataset,
we use Phi-4-mini-instruct \cite{DBLP:journals/corr/abs-2503-01743} and Qwen2.5-7B \cite{DBLP:journals/corr/abs-2412-15115} as the pre-trained models. 
The detailed introduction to these datasets and evaluation metrics will be presented in Appendix \ref{sec:appendix33}.

\textbf{Baselines.}
We compare FISH-Tuning method with the original PEFT method using the same dataset and hyperparameters. We use various datasets, PEFT methods, and pre-trained models to demonstrate that FISH-Tuning is better than the original one. 

We have two ways of comparison: \textbf{The first method} involves fixing the proportion of trainable parameters and comparing FISH-Tuning method with the original PEFT method.
Since FISH-Tuning only fine-tunes a subset of parameters within specific layers, to achieve the same proportion of trainable parameters as the original PEFT method, 
FISH-Tuning method need to fine-tune more layers than the original PEFT method.
\textbf{The second method} involves fixing the number of trainable layers while reducing the proportion of trainable parameters, 
to investigate whether there is similarity between FISH-Tuning method and the original FISH Mask method.



\subsection{Implementation Details}
\label{sec:details}

The initialization weight method of the matrices $B$ and $A$ in LoRA and DoRA follows PiSSA \cite{DBLP:journals/corr/abs-2404-02948}.  For experiments, we did not use any hyper-parameter tuning, nor did we use MLNI trick (use the MLNI checkpoint instead of the pre-trained weights) to enhance the models’ performance. More details about the hyperparameters are available in Table \ref{tab:hyperparameter} in Appendix \ref{sec:appendix3}.

For the experiment fixing the proportion of trainable parameters, we select various layers of the LLMs as trainable parameters to achieve different trainable ratios. 
For most PEFT methods, we select the top 1, 2, 3, and 4 layers to form four groups with different trainable ratios 
(we consider the layers closer to the output layer as the top layers, and the layers closer to the input layer as the bottom layers). 
In FISH-Tuning method, we select the top 5 layers as the parameter set and use the FISH Mask to choose the top-k parameters from this set, 
ensuring the same trainable ratios as the original PEFT method. For $(\text{IA})^3$, we select the top 3, 5, 8, 
and 10 layers to form four groups with different trainable ratios, while in the FISH-Tuning method, 
we use 12 layers because $(\text{IA})^3$ trains fewer parameters compared to other PEFT methods. 

For the experiment fixing the number of trainable layers, we select all layers of the LLMs as trainable parameters.

\subsection{Experimental Results for Fixing the Proportion of Trainable Parameters}
\subsubsection{Baselines}

\begin{table*}[htbp]
    \small
    \begin{tabular*}{\textwidth}{p{2.2cm} c c  @{\hspace{43pt}}  l c c }
        \toprule[1.5pt]
        \textbf{Method} & \textbf{Trainable Parameters}  &  \textbf{Avg} & \textbf{Method} & \textbf{Trainable Parameters}  &  \textbf{Avg}\\

        \midrule
        Original-LoRA & 0.0057\% & 68.45 & Original-PrefixTuning & 0.0439\% &  65.40\\
        LoRA-FISH & 0.0057\% &  \textbf{68.90} & PrefixTuning-FISH & 0.0439\% & \textbf{67.54}\\

        \hdashline
        Original-LoRA & 0.0099\% &  70.42 & Original-PrefixTuning & 0.0864\% & 65.83\\
        LoRA-FISH & 0.0099\% & \textbf{72.25} &  PrefixTuning-FISH & 0.0864\% & \textbf{67.90}\\

        \hdashline
        Original-LoRA & 0.0142\% &  72.08 & Original-PrefixTuning & 0.1289\% & 66.13\\
        LoRA-FISH & 0.0142\% &  \textbf{72.64} & PrefixTuning-FISH & 0.1289\% & \textbf{67.93}\\

        \hdashline
        Original-LoRA & 0.0184\% &  72.11 & Original-PrefixTuning & 0.1713\% & \textbf{69.34}\\
        LoRA-FISH & 0.0184\% & \textbf{72.91} & PrefixTuning-FISH & 0.1713\% &  68.49\\

        \midrule
        Original-DoRA & 0.0078\% & 68.31 & Original-$(\text{IA})^3$ & 0.0142\% &  65.81\\
        DoRA-FISH & 0.0078\% & \textbf{69.45} & $(\text{IA})^3$-FISH & 0.0142\% &  \textbf{67.46}\\

        \hdashline
        Original-DoRA & 0.0142\% &  70.34 &  Original-$(\text{IA})^3$ & 0.0227\% &  67.91 \\
        DoRA-FISH & 0.0142\% &  \textbf{73.09} & $(\text{IA})^3$-FISH & 0.0227\% & \textbf{67.99}\\

        \hdashline
        Original-DoRA & 0.0206\% & 72.59 & Original-$(\text{IA})^3$ & 0.0354\% &  \textbf{67.54}\\
        DoRA-FISH & 0.0206\% & \textbf{73.10} & $(\text{IA})^3$-FISH & 0.0354\% &  67.19\\

        \hdashline
        Original-DoRA & 0.0269\% &  72.27 & Original-$(\text{IA})^3$ & 0.0439\% & 67.64\\
        DoRA-FISH & 0.0269\% & \textbf{73.48} & $(\text{IA})^3$-FISH & 0.0439\% &   \textbf{68.96}\\

        \midrule

        Original-Adapter & 0.1389\% &  67.85 &  Original-UniPELT & 0.0213\% & \textbf{72.04} \\
        Adapter-FISH & 0.1389\% &  \textbf{73.25} & UniPELT-FISH & 0.0213\% & 69.33 \\

        \hdashline
        Original-Adapter & 0.2760\% &  68.91 &  Original-UniPELT & 0.0411\% &  \textbf{72.73} \\
        Adapter-FISH & 0.2760\% & \textbf{73.85} & UniPELT-FISH & 0.0411\% &  70.72\\

        \hdashline
        Original-Adapter & 0.4127\% & 70.36 & Original-UniPELT & 0.0610\% &  \textbf{73.85}\\
        Adapter-FISH & 0.4127\% & \textbf{73.82} & UniPELT-FISH & 0.0610\% &  70.89\\

        \hdashline
        Original-Adapter & 0.5490\% &  72.94 & Original-UniPELT & 0.0808\% & \textbf{72.72} \\
        Adapter-FISH & 0.5490\% & \textbf{73.80} & UniPELT-FISH & 0.0808\% & 70.92\\

        \bottomrule[1.5pt]

    \end{tabular*}
    \caption{
        Performance of different PEFT methods on GLUE dataset. 
        The solid lines separate different PEFT methods, and the dashed lines separate different ratios of
        trainable parameters. In each dashed-line area, the first row represents the original PEFT method, 
        and the second row represents our method.
        The specific details of each dataset result are represented in Appendix \ref{sec:appendix} Table \ref{tab:sumtabwhole}.}
    \label{tab:sumtab}
\end{table*}

We conduct extensive experiments on the GLUE dataset and the BERT model to verify the effectiveness of FISH-Tuning. 
Table \ref{tab:sumtab}  presents the detailed performance of FISH-Tuning and baseline methods on the GLUE benchmark. 
From this table, we can see that integrating the FISH Mask into various PEFT methods consistently improves performance across multiple tasks, except for UniPELT. 
We will discuss why FISH-Tuning performs worse in UniPELT in \nameref{sec:limit}.

\subsubsection{Results with Other Pre-trained Models}

\begin{table}[htbp]
    \small
    \begin{tabular}{lcc}
        \toprule[1.5pt]
        \textbf{Method} & \textbf{Trainable Parameters} &  \textbf{Avg} \\

        \midrule
        Original-LoRA (M) & 0.0033\% &  63.20 \\
        LoRA-FISH & 0.0033\% & \textbf{65.96} \\

        \hdashline
        Original-LoRA (M) & 0.0056\% & 63.50 \\
        LoRA-FISH & 0.0056\% &  \textbf{65.75} \\

        \hdashline
        Original-LoRA (M) & 0.0080\% & 63.97 \\
        LoRA-FISH & 0.0080\% &  \textbf{66.28} \\

        \hdashline
        Original-LoRA (M) & 0.0103\% &  66.17 \\
        LoRA-FISH & 0.0103\% & \textbf{66.81} \\

        \midrule
        Original-LoRA (L) & 0.0007\% & 70.14 \\
        LoRA-FISH & 0.0007\% & \textbf{71.36} \\

        \hdashline
        Original-LoRA (L) & 0.0010\% & 71.36 \\
        LoRA-FISH & 0.0010\% &  \textbf{72.76} \\

        \hdashline
        Original-LoRA (L) & 0.0013\% & 71.88 \\
        LoRA-FISH & 0.0013\% &  \textbf{72.96} \\

        \hdashline
        Original-LoRA (L) & 0.0017\% & 72.87 \\
        LoRA-FISH & 0.0017\% &  \textbf{73.03} \\

        \bottomrule[1.5pt]
    \end{tabular}
    \caption{
        Performance of different PEFT methods on GLUE dataset. (M) means ModernBERT-base. (L) means LLaMA-3.2-1B. 
        The specific details of each dataset result are represented in Appendix \ref{sec:appendix} Table \ref{tab:modernbertandllama3whole}.
    }
    \label{tab:modernbertandllama3}
\end{table}

Apart from the BERT model, we also compare FISH-Tuning with the original PEFT method using other trending pre-trained models, 
such as ModernBERT and LLaMA-3.2-1B. The prompt template for LLaMA-3.2-1B follows Table 1 in \citet{DBLP:journals/corr/abs-2302-10198}. 
The experiment results can be seen in Table \ref{tab:modernbertandllama3}.
These results highlight the generalizability of FISH-Tuning across different pre-trained models, confirming that FISH-Tuning is effective and robust in enhancing the performance of PEFT methods beyond the standard BERT settings.

\subsubsection{Contrastive Study}

\begin{table}[htbp]
    \small
    \begin{tabular}{lcccccccccc}
        \toprule[1.5pt]
        \textbf{Method} & \textbf{Trainable Parameters} & \textbf{Avg} \\

        \midrule
        Original-LoRA & 0.0057\% &  68.45 \\
        LoRA-FISH & 0.0057\% & \textbf{68.90} \\
        LoRA-FISH-rand & 0.0057\%  & 68.74 \\
        LoRA-FISH-rev & 0.0057\% & 66.71 \\

        \hdashline
        Original-LoRA & 0.0099\% &  70.42 \\
        LoRA-FISH & 0.0099\% & \textbf{72.25} \\  
        LoRA-FISH-rand & 0.0099\% &  69.62 \\
        LoRA-FISH-rev & 0.0099\% & 67.93 \\

        \hdashline
        Original-LoRA & 0.0142\% & 72.08 \\
        LoRA-FISH & 0.0142\% & \textbf{72.64} \\
        LoRA-FISH-rand & 0.0142\% & 71.89 \\
        LoRA-FISH-rev & 0.0142\% & 71.86 \\

        \hdashline
        Original-LoRA & 0.0184\%  & 72.11 \\
        LoRA-FISH & 0.0184\%  & 72.91 \\
        LoRA-FISH-rand & 0.0184\% & \textbf{72.92} \\
        LoRA-FISH-rev & 0.0184\% & 72.65 \\

        \bottomrule[1.5pt]
    \end{tabular}
    \caption{
        Performance of different PEFT methods on GLUE dataset. In each dashed-line area, the first row represents the original method, 
        the second row represents our method, the third row represents the method where we randomly select the important parameters without using the FISH Mask, 
        and the fourth row represents the method where we select the important parameters in reverse order compared to the second method.
        The specific details of each dataset result are represented in Appendix \ref{sec:appendix} Table \ref{tab:lorarandandrevwhole}.
    }
    \label{tab:lorarandandrev}
\end{table}

We further investigate the impact of the FISH Mask by conducting a contrastive study on the LoRA framework, as detailed in Table \ref{tab:lorarandandrev}. In this study, we compare the standard LoRA method with three variants: our proposed LoRA-FISH (the standard FISH-Tuning method), a variant where important parameters are selected at random (LoRA-FISH-rand), and another variant where the importance ordering is reversed (LoRA-FISH-rev).
The parameter selection method of LoRA-FISH-rev is described as follows, replacing Equation  \ref{eq:thetaselect}:
\begin{equation}
\label{eq:thetaselectrev}
\theta_{\text{selected}} = \{ \theta_i \ | \ \hat{F}_{\theta_i} \le \text{sort\_reverse}(\hat{F}_{\theta})_k \}
\end{equation}

The results from \ref{tab:lorarandandrev} clearly indicate that the deliberate selection of important parameters using the FISH Mask is key to achieving the performance gains.

\subsection{Experimental Results for Fixing the Number of Trainable Layers}
\subsubsection{Result on Classification Task}

\begin{figure}[htbp]
    \centering
    \centering
    \includegraphics[width=1.0\linewidth]{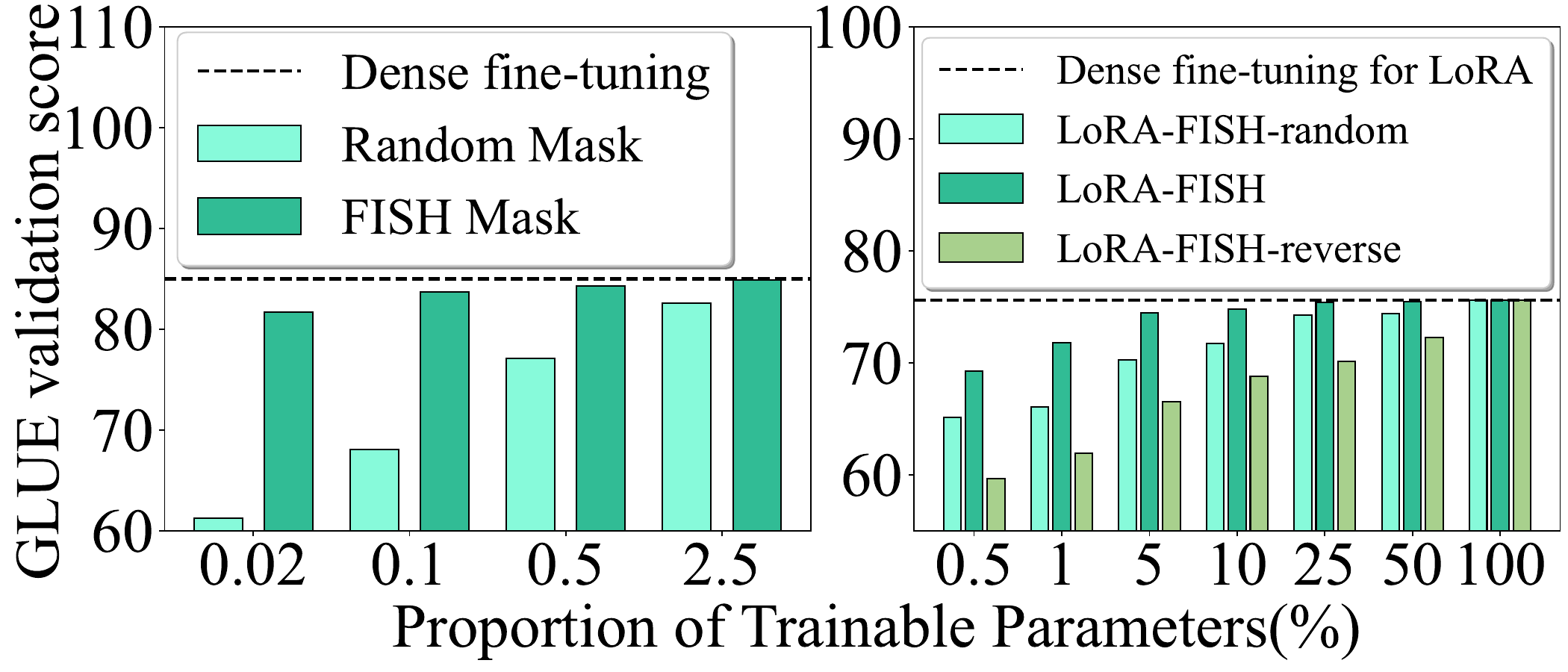}
    \caption{Original FISH Mask method (left) and our FISH-Tuning method (right). 
    The specific details of each dataset result of the right picture are represented in Appendix \ref{sec:appendix} Table \ref{tab:lorarandandrevex2whole}.}
    \label{fig:FISH-Tuningandorigin}
\end{figure}

Fig.~\ref{fig:FISH-Tuningandorigin} presents the detailed performance of FISH-Tuning and baseline methods on the GLUE benchmark. 
In Fig.~\ref{fig:FISH-Tuningandorigin}, the experimental results on the left side of the image are from \cite{DBLP:conf/nips/SungNR21},
and the results on the right side are from our experiments. 
The "proportion of trainable parameters" on the left side represents the ratio of trainable parameters to the total model parameters,
 whereas the "proportion of trainable parameters" on the right side refers to the ratio of trainable parameters to the total parameters within the LoRA component. 
Although FISH Tuning selects the most important subset of parameters from randomly initialized parameters, whereas the original FISH Mask selects the most important subset from the pre-trained model's parameters, both approaches yield similar patterns:
  \textbf{(1)} The parameter selection method of FISH Mask outperforms both Random Mask and Reverse Mask.
  \textbf{(2)} When the proportion value is relatively large, the performance differences among Random Mask, Reverse Mask, and FISH Mask become smaller.


\subsubsection{Result on Generation Task}

\begin{figure}[htbp]
    \centering
    \centering
    \includegraphics[width=1.0\linewidth]{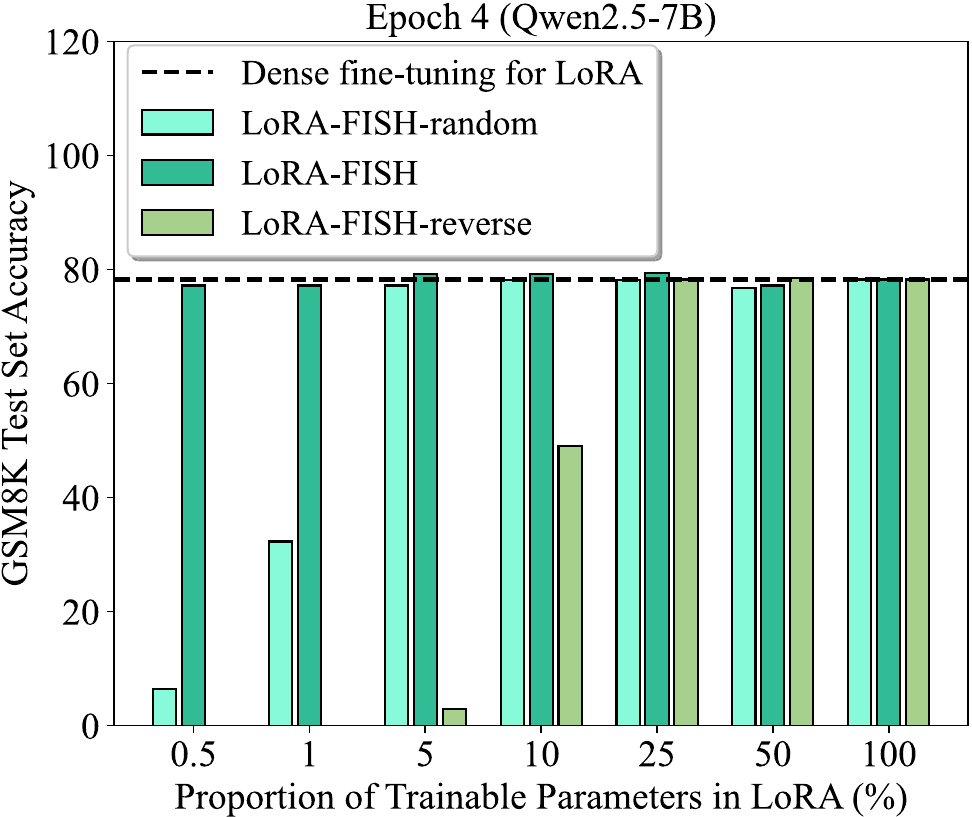}
    \caption{Performance of FISH-Tuning method in different proportion of Trainable Parameters on the GSM8K dataset using Qwen2.5-7B model
     after 4 training epochs.}
    \label{fig:epoqwen4cpy}
\end{figure}

We also conduct experiments on the generation task using the Qwen2.5-7B model. 
Fig.~\ref{fig:epoqwen4cpy} displays the performance results of FISH-Tuning on the GSM8K dataset after completing 4 training epochs. 
As shown in Fig.~\ref{fig:epoqwen4cpy}, when the proportion value is small, the result of FISH Mask is significantly higher than that of Random Mask, 
and the result of Random Mask is clearly higher than that of Reverse Mask. Unlike Fig.~\ref{fig:FISH-Tuningandorigin}, as the proportion value gradually decreases, 
the performance of Random Mask and Reverse Mask drops sharply, while the result of FISH Mask hardly declines at all—sometimes even surpassing Dense fine-tuning. 
This phenomenon indicates that, when training generative tasks with a very small number of trainable parameters, 
selecting parameters for fine-tuning based on Fisher Information can effectively prevent deterioration in model training outcomes.
Pictures for other training epochs are included in the Appendix \ref{sec:appendix4}, Fig.~\ref{fig:epoqwenwhole}.

\begin{figure}[htbp]
    \centering
    \centering
    \includegraphics[width=1.0\linewidth]{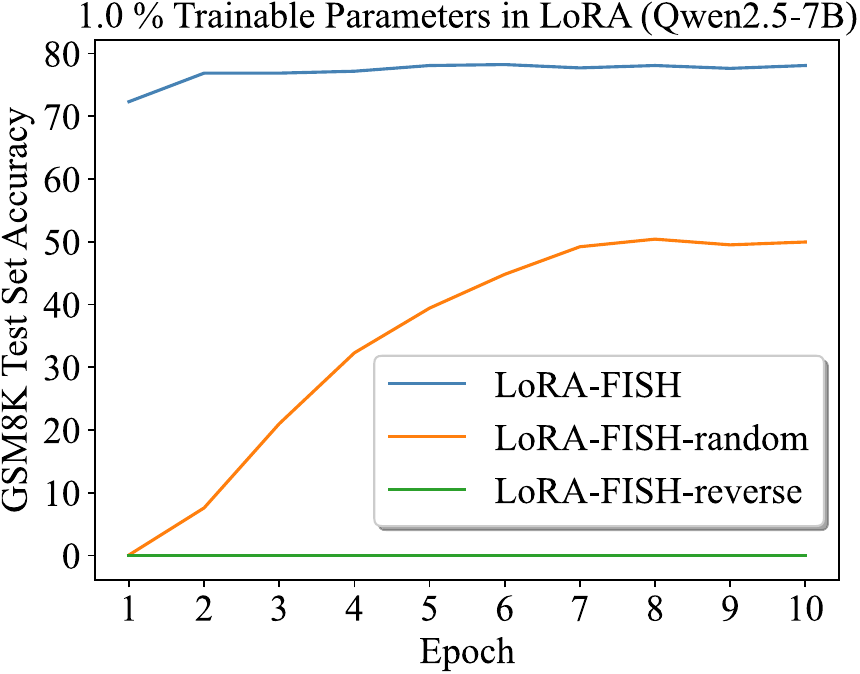}
    \caption{LoRA-FISH converges faster compared to LoRA-FISH-random and LoRA-FISH-reverse.}
    \label{fig:proqwen1.0cpy}
\end{figure}

Fig.~\ref{fig:proqwen1.0cpy} shows the performance of FISH-Tuning across different training epochs when the proportion value is set to 1\%. 
From Fig.~\ref{fig:proqwen1.0cpy}, we observe that FISH Mask nearly converges after just 1 training epoch, while Random Mask approaches convergence only after 7 epochs. 
In contrast, Reverse Mask maintains an accuracy rate of zero throughout. By the 10-th training epoch, 
the result of FISH Mask is significantly better than that of Random Mask, and Random Mask is significantly better than Reverse Mask. 
The results presented in Fig.~\ref{fig:proqwen1.0cpy} further confirm the importance of using Fisher Information to select parameters for fine-tuning. 
Pictures for other proportion values are provided in the Appendix \ref{sec:appendix4}, Fig.~\ref{fig:proqwenwhole}.

In addition to Qwen2.5-7B, similar experiments were conducted on Phi-4-mini-instruct, where comparable results are observed. 
The experimental results for Phi-4-mini-instruct are detailed in the Appendix \ref{sec:appendix5}, Fig.~\ref{fig:epophiwhole} and Fig.~\ref{fig:prophiwhole}.

\subsubsection{Comparison with LoRI}

\begin{table}[htbp]
    \small
    \begin{tabular}{l ccccc}
        \toprule[1.5pt]
        \multirow{2}{*}{\textbf{Method}} & \multicolumn{5}{c}{\textbf{Trainable parameters in LoRA (\textperthousand)}} \\
         & \textbf{0.018} & \textbf{0.036} & \textbf{0.18} & \textbf{0.36} & \textbf{0.9} \\
        \midrule
        LoRI & 78.20 &  \textbf{79.14}  &  \textbf{79.53} &  78.13 &  77.50  \\
        FISHTuning & \textbf{78.24} & 76.88  & 78.01 & \textbf{78.77} & \textbf{78.47}  \\
        \bottomrule[1.5pt]
    \end{tabular}
    \caption{The performance of LoRI and our FISH-Tuning method in the same proportion of trainable parameters on Qwen2.5-7B.}
    \label{tab:lori}
\end{table}

We compared our method with the LoRI \cite{zhang2025lori} approach, and the comparison results are shown in Table \ref{tab:lori}. 
We set the same hyperparameters to ensure fairness. As illustrated in Table \ref{tab:lori}, 
our FISH-Tuning method outperforms LoRI under three proportions of trainable parameter ratios, 
while it performs worse than LoRI under two proportions. Overall, these experimental results validate 
the effectiveness of FISH-Tuning in balancing model performance and parameter selection, 
demonstrating its potential as a novel fine-tuning approach.

\subsubsection{Resource Consumption Analysis}

We compare the changes in resource consumption caused by adding FISH-Tuning to the original PEFT methods. 
Table \ref{tab:resourceanaly}  presents the resource consumption when training a BERT model on the RTE task for 1 epoch, 
under the condition of fixing the number of trainable layers. 
The first row in Table \ref{tab:resourceanaly} shows the runtime and GPU memory usage when using the original LoRA method for training, 
and the other rows show the runtime and GPU memory usage when fine-tuning part of the parameters in LoRA component after using FISH-Tuning.
As the proportion of trainable parameters decreases, 
the required runtime decreases while GPU memory slightly increases. 
The runtime reduction results from fewer parameters requiring fine-tuning within the model, 
while the increase in GPU memory is due to additional matrices introduced by the FISH Mask consuming memory. 
In theory, if GPUs were to provide hardware-level support for sparse matrix parameter updates, 
the GPU memory consumption would also decrease. Incorporating FISH-Tuning does not affect the time 
or GPU memory consumed during inference since the FISH Mask is only used during training.

\begin{table}[htbp]
    \small
    \begin{tabular}{l|c  c c c }
        \toprule[1.5pt]
        \textbf{Method} & \textbf{Ratio1} & \textbf{Ratio2}  & \textbf{Time} & \textbf{GPU} \\

        \midrule
        Origin-LoRA & 0.8116\% & 100\% & 38.0208 & 6180 \\

        LoRA-FISH & 0.4058\% & 50\% & 37.6212 & 6650	 \\

        LoRA-FISH & 0.2029\% & 25\% &  37.4502 &	6650  \\

        LoRA-FISH & 0.0812\% & 10\% &  37.7291	& 6650  \\

        LoRA-FISH & 0.0406\% & 5\% &  	37.0493	& 6650	\\

        LoRA-FISH & 0.0081\% & 1\% &  37.1929	&  6650	 \\

        LoRA-FISH & 0.0041\% & 0.5\% &  36.6827	& 6650  \\

        \bottomrule[1.5pt]
    \end{tabular}
    \caption{Resource consumption of FISH-Tuning method in different proportion of Trainable Parameters. 
    The "Ratio1" column represents the ratio of trainable parameters to the total number of parameters, 
    the "Ratio2" column represents the ratio of trainable parameters to the total number of parameters in LoRA component,
    the "Time" column represents the training time in seconds, and the "GPU" column represents the GPU memory consumption in MB.}
    \label{tab:resourceanaly}
\end{table}

\section{Conclusion}
In this paper, we propose FISH-Tuning, a new method that uses the selective PEFT method in the Addition-based and Reparameterization-based PEFT methods. 
More concretely, we integrate the FISH Mask into LoRA, DoRA, Adapter, Prefix-Tuning, and $(\text{IA})^3$. 
With the same ratio of trainable parameters, our method outperforms the original PEFT method most of the time. 
With the same number of trainable layers and ranks, our method achieves experimental results similar to those of the original FISH Mask method.
In future work, we will explore our approach in Computer Vision, Multi-modality, and Quantization.

\section*{Limitations}
\label{sec:limit}
There are still some questions with FISH-Tuning not addressed in this paper:

1) FISH-Tuning does not perform well within the UniPELT framework. We hypothesize that the reason for this phenomenon is that UniPELT uses a gating value, ${\mathcal{G}_m} \in (0, 1)$. Each individual PEFT module might possess "emergent abilities" similar to causal language models \cite{DBLP:journals/tmlr/WeiTBRZBYBZMCHVLDF22}. While an individual PEFT module can perform well using FISH-Tuning, when using UniPELT, the weights of each PEFT module are multiplied by a gating value, ${\mathcal{G}_m} \in (0, 1)$. This gating value may "dilute" the individual PEFT module's "emergent abilities". This explanation is merely our hypothesis and requires further experimental validation.

2) We use the LoRA method as an example. To achieve the same ratio of trainable parameters in FISH-Tuning, we can select different trainable layers and LoRA ranks. Tables \ref{tab:loradifferentlayer} and \ref{tab:loradifferentrank} in Appendix \ref{sec:appendix2} demonstrate our experiments on this. According to the results, we have not found any obvious pattern for determining how to select the suitable ranks and layers to achieve the best result. 

Nevertheless, we are excited to see the huge potential of FISH-Tuning already demonstrated in existing experiments and look forward to more tests and suggestions from the community.
In future work, we will try to incorporate other Selection-based PEFT methods into Addition-based PEFT methods 
and Reparameterization-based PEFT methods and compare them with our FISH-Tuning method. 

\section*{Ethics Statement}
This work aims to contribute to the advancement of Machine Learning. We encourage ethical and responsible application of our results to prevent societal harm. To ensure transparency and reproducibility, and to promote trust and integrity in Machine Learning research, we will release our code and methods publicly. 

\section*{Acknowledgments}
We use generative AI tools only to assist with the language of the paper. We only use them to check for grammatical issues. All new ideas and new text are written by ourselves.

\bibliography{custom}


\appendix

\section{Loss on Evaluation Datasets}
\label{sec:example}

\begin{figure}[htbp]
    \centering
    \centering
    \includegraphics[width=1.0\linewidth]{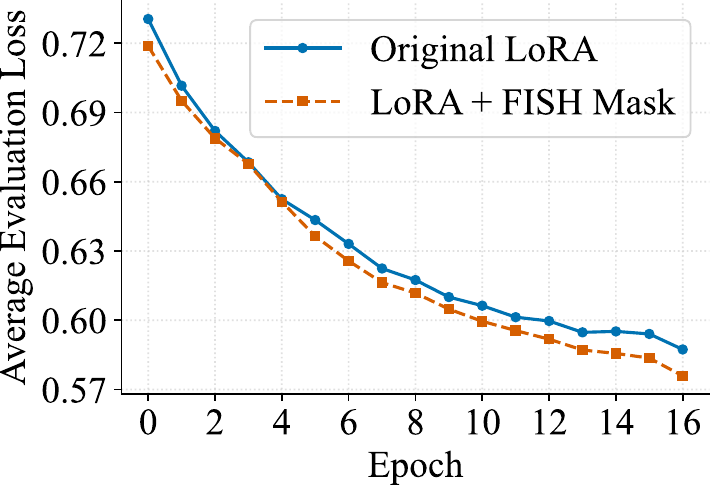}
    \caption{Original LoRA method loss (Blue line) and the LoRA-FISH method loss (Orange line).
    The x-axis represents the training epochs, and the y-axis denotes the average evaluation loss scores across six datasets. }
    \label{fig:loss}
\end{figure}

We conduct an experiment on the loss scores using the evaluation dataset. The loss values are taken from the first solid-line area in Table \ref{tab:sumtab}. 
We visualize the loss value curves for the original PEFT method and FISH-Tuning method in Fig. \ref{fig:loss}. 
From this figure, we observe that all points on the FISH-Tuning curve are lower than those on the original PEFT method curve. 
This result demonstrates that the FISH-Tuning method converges faster than the original PEFT method. 
It also explains why FISH-Tuning method achieves better results in Table \ref{tab:sumtab}.

\section{Tables of Precision Results}
\label{sec:appendix}

This section will present the model performance results on the GLUE dataset mentioned in the main text. 
Due to space limitations, the tables in the main text only list the average scores of each model on the GLUE dataset,
whereas this section will further provide the detailed performance metrics of each model on individual subtasks of GLUE.

Table \ref{tab:sumtabwhole} and  Table \ref{tab:modernbertandllama3whole} present the detailed performance of FISH-Tuning and base-line methods on the GLUE benchmark.
The pre-trained model of Table \ref{tab:sumtabwhole} is bert-base-cased. 
The pre-trained models of Table \ref{tab:modernbertandllama3whole} are ModernBERT-base and LLaMA-3.2-1B.

Table \ref{tab:lorarandandrevwhole} compare the standard LoRA method with three variants: our proposed LoRA-FISH (the standard FISH-Tuning method), 
a variant where important parameters are selected at random(LoRA-FISH-rand), and another variant where the importance ordering is reversed (LoRA-FISH-rev). 
The pre-trained model is bert-base-cased.

\begin{table*}[htbp]
    \small
    \begin{tabular*}{\textwidth}{p{3cm}cccccccccc}
        \toprule[1.5pt]
        \textbf{Method} & \textbf{Trainable Parameters} & \textbf{CoLA} & \textbf{MRPC} & \textbf{RTE} & \textbf{SST-2} & \textbf{STS-B} & \textbf{WNLI} & \textbf{Avg} \\

        \midrule
        Original-LoRA & 0.0057\% & 43.27 & \textbf{80.94} & 58.48 & 89.56 & 84.94 & 53.52 & 68.45 \\
        LoRA-FISH & 0.0057\% & \textbf{44.28} & 80.25 & 58.48 & \textbf{90.48} & \textbf{86.41} & 53.52 & \textbf{68.90} \\

        \hdashline
        Original-LoRA & 0.0099\% & 48.02 & 82.17 & 62.82 & 89.91 & 86.06 & 53.52 & 70.42 \\
        LoRA-FISH & 0.0099\% & \textbf{51.21} & \textbf{85.74} & \textbf{66.06} & \textbf{90.14} & \textbf{86.84} & 53.52 & \textbf{72.25} \\

        \hdashline
        Original-LoRA & 0.0142\% & 51.87 & 84.35 & 64.98 & \textbf{91.17} & 86.60 & 53.52 & 72.08 \\
        LoRA-FISH & 0.0142\% & \textbf{53.58} & \textbf{85.56} & \textbf{65.70} & 90.71 & \textbf{86.79} & 53.52 & \textbf{72.64} \\

        \hdashline
        Original-LoRA & 0.0184\% & 54.96 & 82.41 & 64.62 & \textbf{90.14} & \textbf{87.03} & 53.52 & 72.11 \\
        LoRA-FISH & 0.0184\% & \textbf{55.96} & \textbf{84.18} & \textbf{67.51} & 89.45 & 86.81 & 53.52 & \textbf{72.91} \\

        \midrule
        Original-DoRA & 0.0078\% & 42.70 & \textbf{80.45} & 58.48 & 89.68 & 85.04 & 53.52 & 68.31 \\
        DoRA-FISH & 0.0078\% & \textbf{46.36} & 80.30 & 58.48 & \textbf{91.17} & \textbf{86.84} & 53.52 & \textbf{69.45} \\

        \hdashline
        Original-DoRA & 0.0142\% & 47.83 & 80.49 & 63.90 & 90.14 & 86.14 & 53.52 & 70.34 \\
        DoRA-FISH & 0.0142\% & \textbf{54.14} & \textbf{87.47} & \textbf{66.43} & 90.14 & \textbf{86.84} & 53.52 & \textbf{73.09} \\

        \hdashline
        Original-DoRA & 0.0206\% & 54.54 & 84.00 & 65.70 & \textbf{91.06} & 86.73 & 53.52 & 72.59 \\
        DoRA-FISH & 0.0206\% & \textbf{55.89} & \textbf{85.63} & \textbf{66.79} & 89.91 & \textbf{86.86} & 53.52 & \textbf{73.10} \\

        \hdashline
        Original-DoRA & 0.0269\% & 54.13 & 82.58 & 66.06 & \textbf{90.25} & 87.07 & 53.52 & 72.27 \\
        DoRA-FISH & 0.0269\% & \textbf{56.00} & \textbf{86.70} & \textbf{67.51} & 89.91 & \textbf{87.21} & 53.52 & \textbf{73.48} \\

        \midrule

        Original-Adapter & 0.1389\% & 41.53 & 78.54 & 62.82 & 85.32 & 82.54 & 56.34 & 67.85 \\
        Adapter-FISH & 0.1389\% & \textbf{52.27} & \textbf{87.36} & \textbf{64.26} & \textbf{90.60} & \textbf{87.26} & \textbf{57.75} & \textbf{73.25} \\

        \hdashline
        Original-Adapter & 0.2760\% & 47.05 & 83.58 & 62.09 & \textbf{90.60} & 86.47 & 43.66 & 68.91 \\
        Adapter-FISH & 0.2760\% & \textbf{53.73} & \textbf{88.91} & \textbf{64.62} & 90.48 & \textbf{87.62} & \textbf{57.75} & \textbf{73.85} \\

        \hdashline
        Original-Adapter & 0.4127\% & 48.44 & 85.05 & 61.73 & 90.71 & 86.94 & 49.30 & 70.36 \\
        Adapter-FISH & 0.4127\% & \textbf{52.64} & \textbf{87.76} & \textbf{65.70} & \textbf{91.28} & \textbf{87.76} & \textbf{57.75} & \textbf{73.82} \\

        \hdashline
        Original-Adapter & 0.5490\% & 51.85 & 87.97 & \textbf{64.62} & 90.48 & 87.76 & 54.93 & 72.94 \\
        Adapter-FISH & 0.5490\% & \textbf{53.13} & \textbf{88.19} & 64.26 & \textbf{91.51} & \textbf{87.98} & \textbf{57.75} & \textbf{73.80} \\

        \midrule
        Original-PrefixTuning & 0.0439\% & 34.03 & 76.57 & 59.57 & 83.83 & 80.65 & 57.75 & 65.40 \\
        PrefixTuning-FISH & 0.0439\% & \textbf{38.17} & \textbf{76.60} & \textbf{60.29} & \textbf{88.07} & \textbf{82.92} & \textbf{59.15} & \textbf{67.54} \\

        \hdashline
        Original-PrefixTuning & 0.0864\% & 32.10 & 76.94 & \textbf{61.37} & 85.78 & 82.48 & 56.34 & 65.83 \\
        PrefixTuning-FISH & 0.0864\% & \textbf{39.22} & \textbf{77.15} & 60.29 & \textbf{87.61} & \textbf{83.94} & \textbf{59.15} & \textbf{67.90} \\

        \hdashline
        Original-PrefixTuning & 0.1289\% & 35.05 & 74.96 & \textbf{60.65} & 87.50 & 83.72 & 54.93 & 66.13 \\
        PrefixTuning-FISH & 0.1289\% & \textbf{38.91} & \textbf{77.13} & 60.29 & \textbf{87.61} & \textbf{84.47} & \textbf{59.15} & \textbf{67.93} \\

        \hdashline
        Original-PrefixTuning & 0.1713\% & 35.80 & \textbf{77.87} & \textbf{65.34} & 88.19 & 84.04 & \textbf{64.79} & \textbf{69.34} \\
        PrefixTuning-FISH & 0.1713\% & \textbf{40.94} & 77.13 & 60.29 & \textbf{88.76} & \textbf{84.69} & 59.15 & 68.49 \\

        \midrule
        Original-$(\text{IA})^3$ & 0.0142\% & 34.11 & 76.42 & 64.62 & 87.16 & 84.68 & 47.89 & 65.81 \\
        $(\text{IA})^3$-FISH & 0.0142\% & \textbf{38.54} & \textbf{78.03} & 64.62 & \textbf{89.11} & \textbf{86.60} & 47.89 & \textbf{67.46} \\

        \hdashline
        Original-$(\text{IA})^3$ & 0.0227\% & \textbf{40.64} & 78.68 & \textbf{65.70} & 88.65 & 85.93 & 47.89 & 67.91 \\
        $(\text{IA})^3$-FISH & 0.0227\% & 38.47 & \textbf{81.28} & 64.26 & \textbf{88.99} & \textbf{87.04} & 47.89 & \textbf{67.99} \\

        \hdashline
        Original-$(\text{IA})^3$ & 0.0354\% & \textbf{40.45} & 76.73 & \textbf{64.98} & \textbf{89.68} & 86.94 & 46.48 & \textbf{67.54} \\
        $(\text{IA})^3$-FISH & 0.0354\% & 37.99 & \textbf{80.06} & 62.09 & 89.11 & \textbf{87.41} & 46.48 & 67.19 \\

        \hdashline
        Original-$(\text{IA})^3$ & 0.0439\% & 39.59 & \textbf{81.47} & 61.73 & 89.11 & 87.44 & 46.48 & 67.64 \\
        $(\text{IA})^3$-FISH & 0.0439\% & \textbf{46.74} & 81.26 & \textbf{62.45} & \textbf{89.22} & \textbf{87.59} & 46.48 & \textbf{68.96} \\

        \midrule
        Original-UniPELT & 0.0213\% & 47.25 & \textbf{81.76} & 60.65 & \textbf{90.71} & 85.65 & \textbf{66.20} & \textbf{72.04} \\
        UniPELT-FISH & 0.0213\% & \textbf{49.72} & 81.28 & \textbf{64.26} & 90.14 & \textbf{86.94} & 43.66 & 69.33 \\

        \hdashline
        Original-UniPELT & 0.0411\% & 53.65 & \textbf{84.42} & \textbf{64.62} & 90.48 & 86.85 & \textbf{56.34} & \textbf{72.73} \\
        UniPELT-FISH & 0.0411\% & \textbf{55.70} & 81.95 & 63.90 & \textbf{91.74} & \textbf{87.39} & 43.66 & 70.72 \\

        \hdashline
        Original-UniPELT & 0.0610\% & \textbf{55.89} & 87.29 & \textbf{64.26} & 90.94 & 86.96 & \textbf{57.75} & \textbf{73.85} \\
        UniPELT-FISH & 0.0610\% & 51.66 & \textbf{87.38} & 63.90 & \textbf{91.28} & \textbf{87.45} & 43.66 & 70.89 \\

        \hdashline
        Original-UniPELT & 0.0808\% & \textbf{52.12} & \textbf{88.08} & \textbf{65.70} & \textbf{90.94} & 87.36 & \textbf{52.11} & \textbf{72.72} \\
        UniPELT-FISH & 0.0808\% & 52.09 & 87.80 & 63.90 & 90.60 & \textbf{87.45} & 43.66 & 70.92 \\

        \bottomrule[1.5pt]

    \end{tabular*}
    \caption{Performance of different PEFT methods on GLUE dataset. The solid lines separate different PEFT methods, and the dashed lines separate different ratios of trainable parameters. In each dashed-line area, the first row represents the original method, and the second row represents our method.}
    \label{tab:sumtabwhole}
\end{table*}

\begin{table*}[htbp]
    \small
    \begin{tabular*}{\textwidth}{p{3cm}cccccccccc}
        \toprule[1.5pt]
        \textbf{Method} & \textbf{Trainable Parameters} & \textbf{CoLA} & \textbf{MRPC} & \textbf{RTE} & \textbf{SST-2} & \textbf{STS-B} & \textbf{WNLI} & \textbf{Avg} \\

        \midrule
        Original-LoRA (M) & 0.0033\% & 33.93 & 79.14 & 57.04 & 83.37 & 79.24 & 46.48 & 63.20 \\
        LoRA-FISH & 0.0033\% & \textbf{36.60} & \textbf{80.14} & \textbf{57.76} & \textbf{86.24} & \textbf{81.52} & \textbf{53.52} & \textbf{65.96} \\

        \hdashline
        Original-LoRA (M) & 0.0056\% & 33.75 & \textbf{79.82} & 57.76 & 83.60 & 79.57 & 46.48 & 63.50 \\
        LoRA-FISH & 0.0056\% & \textbf{35.25} & 79.44 & \textbf{58.12} & \textbf{86.24} & \textbf{81.95} & \textbf{53.52} & \textbf{65.75} \\

        \hdashline
        Original-LoRA (M) & 0.0080\% & \textbf{37.48} & 79.18 & 57.40 & 84.98 & 79.73 & 45.07 & 63.97 \\
        LoRA-FISH & 0.0080\% & 36.44 & \textbf{79.97} & \textbf{58.48} & \textbf{86.58} & \textbf{82.66} & \textbf{53.52} & \textbf{66.28} \\

        \hdashline
        Original-LoRA (M) & 0.0103\% & \textbf{41.72} & \textbf{79.44} & 57.76 & 84.86 & 81.13 & 52.11 & 66.17 \\
        LoRA-FISH & 0.0103\% & 40.04 & 79.10 & \textbf{58.48} & \textbf{87.16} & \textbf{82.58} & \textbf{53.52} & \textbf{66.81} \\

        \midrule
        Original-LoRA (L) & 0.0007\% & 46.80 & 77.31 & 68.59 & 89.68 & 80.71 & 57.75 & 70.14 \\
        LoRA-FISH & 0.0007\% & \textbf{47.07} & \textbf{79.67} & \textbf{70.04} & \textbf{90.94} & \textbf{82.70} & 57.75 & \textbf{71.36} \\

        \hdashline
        Original-LoRA (L) & 0.0010\% & 48.96 & 78.09 & 71.48 & 90.83 & 82.49 & 56.34 & 71.36 \\
        LoRA-FISH & 0.0010\% & \textbf{52.67} & \textbf{80.24} & \textbf{72.20} & \textbf{91.17} & \textbf{83.93} & 56.34 & \textbf{72.76} \\

        \hdashline
        Original-LoRA (L) & 0.0013\% & 49.72 & 78.19 & 70.76 & 91.28 & 83.60 & \textbf{57.75} & 71.88 \\
        LoRA-FISH & 0.0013\% & \textbf{52.16} & \textbf{80.43} & \textbf{72.20} & \textbf{91.86} & \textbf{84.77} & 56.34 & \textbf{72.96} \\

        \hdashline
        Original-LoRA (L) & 0.0017\% & \textbf{54.03} & 79.24 & \textbf{72.92} & \textbf{91.06} & 83.61 & 56.34 & 72.87 \\
        LoRA-FISH & 0.0017\% & 53.84 & \textbf{80.45} & 72.56 & 90.71 & \textbf{84.29} & 56.34 & \textbf{73.03} \\

        \bottomrule[1.5pt]
    \end{tabular*}
    \caption{Performance of different PEFT methods on GLUE dataset. (M) means ModernBERT-base. (L) means LLaMA-3.2-1B.}
    \label{tab:modernbertandllama3whole}
\end{table*}

\begin{table*}[htbp]
    \small
    \begin{tabular*}{\textwidth}{p{3cm}cccccccccc}
        \toprule[1.5pt]
        \textbf{Method} & \textbf{Trainable Parameters} & \textbf{CoLA} & \textbf{MRPC} & \textbf{RTE} & \textbf{SST-2} & \textbf{STS-B} & \textbf{WNLI} & \textbf{Avg} \\

        \midrule
        Original-LoRA & 0.0057\% & 43.27 & \textbf{80.94} & 58.48 & 89.56 & 84.94 & 53.52 & 68.45 \\
        LoRA-FISH & 0.0057\% & 44.28 & 80.25 & 58.48 & \textbf{90.48} & 86.41 & 53.52 & \textbf{68.90} \\
        LoRA-FISH-rand & 0.0057\% & \textbf{46.18} & 78.38 & 58.12 & 89.56 & \textbf{86.69} & 53.52 & 68.74 \\
        LoRA-FISH-rev & 0.0057\% & 38.91 & 74.80 & 58.12 & 88.99 & 85.91 & 53.52 & 66.71 \\

        \hdashline
        Original-LoRA & 0.0099\% & 48.02 & 82.17 & 62.82 & 89.91 & 86.06 & 53.52 & 70.42 \\
        LoRA-FISH & 0.0099\% & \textbf{51.21} & \textbf{85.74} & \textbf{66.06} & \textbf{90.14} & 86.84 & 53.52 & \textbf{72.25} \\  
        LoRA-FISH-rand & 0.0099\% & 48.67 & 81.25 & 58.12 & 89.22 & \textbf{86.92} & 53.52 & 69.62 \\
        LoRA-FISH-rev & 0.0099\% & 44.53 & 74.80 & 58.12 & 89.91 & 86.68 & 53.52 & 67.93 \\

        \hdashline
        Original-LoRA & 0.0142\% & 51.87 & 84.35 & 64.98 & 91.17 & 86.60 & 53.52 & 72.08 \\
        LoRA-FISH & 0.0142\% & \textbf{53.58} & \textbf{85.56} & 65.70 & 90.71 & 86.79 & 53.52 & \textbf{72.64} \\
        LoRA-FISH-rand & 0.0142\% & 52.07 & 82.24 & 66.43 & 90.14 & 86.95 & 53.52 & 71.89 \\
        LoRA-FISH-rev & 0.0142\% & 48.68 & 84.03 & 66.43 & 91.17 & \textbf{87.33} & 53.52 & 71.86 \\

        \hdashline
        Original-LoRA & 0.0184\% & 54.96 & 82.41 & 64.62 & 90.14 & 87.03 & 53.52 & 72.11 \\
        LoRA-FISH & 0.0184\% & \textbf{55.96} & 84.18 & 67.51 & 89.45 & 86.81 & 53.52 & 72.91 \\
        LoRA-FISH-rand & 0.0184\% & 54.59 & \textbf{85.49} & 66.79 & 89.91 & \textbf{87.23} & 53.52 & \textbf{72.92} \\
        LoRA-FISH-rev & 0.0184\% & 54.10 & 83.31 & 67.51 & \textbf{90.25} & 87.20 & 53.52 & 72.65 \\

        \bottomrule[1.5pt]
    \end{tabular*}
    \caption{Performance of different PEFT methods on GLUE dataset. In each dashed-line area, the first row represents the original method, the second row represents our method, the third row represents the method where we randomly select the important parameters without using the FISH Mask, and the fourth row represents the method where we select the important parameters in reverse order compared to the second method.}
    \label{tab:lorarandandrevwhole}
\end{table*}

\begin{table*}[htbp]
    \small
    \begin{tabular*}{\textwidth}{p{3cm} @{\hspace{10pt}} c @{\hspace{20pt}}  c @{\hspace{20pt}} ccccccc}
        \toprule[1.5pt]
        \textbf{Method} & \textbf{Ratio1} & \textbf{Ratio2}  & \textbf{CoLA} & \textbf{MRPC} & \textbf{RTE} & \textbf{SST-2} & \textbf{STS-B} & \textbf{WNLI} & \textbf{Avg} \\

        \midrule
        Original-LoRA & 0.8116\% & 100\% & 60.85 & 88.26 & 70.40 & 91.86 & 88.77 & 53.52 & 75.61 \\

        \hdashline
        LoRA-FISH & 0.4058\% & 50\% & \textbf{60.08} & \textbf{87.71} & 	\textbf{70.40} & 90.94 & \textbf{88.67} & \textbf{54.93} & \textbf{75.45} \\
        LoRA-FISH-rand & 0.4058\% & 50\% & 57.54 & 86.36	& 69.31	& 91.28	& 88.29	& 53.52	& 74.39  \\
        LoRA-FISH-rev & 0.4058\% & 50\% & 58.56	& 80.35	& 61.37 &	\textbf{91.97} &	87.68 &	53.52 &	72.24 \\

        \hdashline
        LoRA-FISH & 0.2029\% & 25\% &  \textbf{58.30} &	\textbf{86.72} &	\textbf{72.92} &	91.06 & 	\textbf{88.44} &	\textbf{54.93}	& \textbf{75.39} \\
        LoRA-FISH-rand & 0.2029\% & 25\% & 	56.29	& 85.45	& 71.12	& 91.17	& 87.99 &	53.52	& 74.26 \\
        LoRA-FISH-rev & 0.2029\% & 25\% &  51.62	& 78.27	& 58.84	& \textbf{91.40}	& 87.00	& 53.52	& 70.11  \\

        \hdashline
        LoRA-FISH & 0.0812\% & 10\% &  \textbf{55.51}	& \textbf{87.40}	& \textbf{72.56}	& 90.25	& \textbf{88.30}	& \textbf{54.93}	& \textbf{74.83}  \\
        LoRA-FISH-rand & 0.0812\% & 10\% &  52.89	&  81.64	&  63.18	&  \textbf{91.17}	&  87.91	&  53.52	&  71.72  \\
        LoRA-FISH-rev & 0.0812\% & 10\% &  49.95	& 74.80	& 58.12	& 90.14	& 86.27	& 53.52	& 68.80  \\

        \hdashline
        LoRA-FISH & 0.0406\% & 5\% &  	\textbf{54.84}	& \textbf{86.04}	& \textbf{72.56}	& 90.37	& \textbf{88.21}	& \textbf{54.93}	& \textbf{74.49}  \\
        LoRA-FISH-rand & 0.0406\% & 5\% &  49.11	& 80.90	& 58.84	& \textbf{91.28} &	87.84	& 53.52 &	70.25  \\
        LoRA-FISH-rev & 0.0406\% & 5\% &  	39.18	& 74.80	& 58.12	& 88.42	& 85.00	& 53.52	& 66.51 \\

        \hdashline
        LoRA-FISH & 0.0081\% & 1\% &  \textbf{46.64}	&  \textbf{85.80}	& \textbf{67.51} & 	89.22	& \textbf{88.14}	& \textbf{53.52}	& \textbf{71.81}  \\
        LoRA-FISH-rand & 0.0081\% & 1\% & 34.44	& 74.80	& 58.48	& \textbf{89.68}	& 85.46	& 53.52	& 66.07  \\
        LoRA-FISH-rev & 0.0081\% & 1\% &  25.12	&  74.80	& 58.12	& 83.83	& 76.06	& 53.52	& 61.91  \\

        \hdashline
        LoRA-FISH & 0.0041\% & 0.5\% &  \textbf{40.71}	& \textbf{81.71} & \textbf{63.54}	& \textbf{88.07} & 	\textbf{88.09}	& \textbf{53.52}	& \textbf{69.27}  \\
        LoRA-FISH-rand & 0.0041\% & 0.5\% & 34.55	 & 74.80	 & 58.12	 & 87.61	 & 82.19  & 	53.52	 & 65.13  \\
        LoRA-FISH-rev & 0.0041\% & 0.5\% &  	18.81	& 74.80	& 58.12	& 81.77	& 71.12	& 53.52	& 59.69        \\

        \bottomrule[1.5pt]
    \end{tabular*}
    \caption{Performance of different PEFT methods on GLUE dataset. The "Ratio1" column represents the ratio of trainable parameters to the total number of parameters, 
    and the "Ratio2" column represents the ratio of trainable parameters to the total number of parameters in LoRA component. }
    \label{tab:lorarandandrevex2whole}
\end{table*}

\section{Different Layers and Ranks for LoRA}
\label{sec:appendix2}

\begin{table*}[htbp]
    \small
    \begin{tabular*}{\textwidth}{p{4cm}cccccccccc}
        \toprule[1.5pt]
        \textbf{Method} & \textbf{Trainable Parameters} & \textbf{CoLA} & \textbf{MRPC} & \textbf{RTE} & \textbf{STS-B} & \textbf{WNLI} & \textbf{Avg} \\

        \midrule
        Original-LoRA & 0.0057\% & 43.27 & 80.94 & 58.48 & 84.94 & 53.52 & 64.23 \\
        LoRA-FISH-rk1-lay5 & 0.0057\% & 44.28 & 80.25 & 58.48 & 86.41 & 53.52 & 64.59 \\
        LoRA-FISH-rk1-lay6 & 0.0057\% & 44.19 & \textbf{84.73} & 58.48 & 86.93 & 53.52 & 65.57 \\
        LoRA-FISH-rk1-lay8 & 0.0057\% & 46.92 & 84.66 & 58.48 & \textbf{87.12} & 53.52 & \textbf{66.14} \\
        LoRA-FISH-rk1-lay10 & 0.0057\% & 46.05 & 80.01 & 58.48 & 86.99 & 53.52 & 65.01 \\
        LoRA-FISH-rk1-lay12 & 0.0057\% & \textbf{46.99} & 84.53 & 58.48 & 87.11 & 53.52 & 66.13 \\

        \hdashline
        Original-LoRA & 0.0099\% & 48.02 & 82.17 & 62.82 & 86.06 & 53.52 & 66.52 \\
        LoRA-FISH-rk1-lay5 & 0.0099\% & 51.21 & \textbf{85.74} & \textbf{66.06} & 86.84 & 53.52 & \textbf{68.67} \\
        LoRA-FISH-rk1-lay6 & 0.0099\% & 51.23 & 85.43 & 64.98 & 87.06 & 53.52 & 68.45 \\
        LoRA-FISH-rk1-lay8 & 0.0099\% & 52.61 & 85.71 & 58.48 & 87.23 & 53.52 & 67.51 \\
        LoRA-FISH-rk1-lay10 & 0.0099\% & \textbf{53.41} & 80.23 & 65.70 & 87.37 & 53.52 & 68.05 \\
        LoRA-FISH-rk1-lay12 & 0.0099\% & 51.03 & 84.99 & 58.48 & \textbf{87.44} & 53.52 & 67.09 \\

        \hdashline
        Original-LoRA & 0.0142\% & 51.87 & 84.35 & 64.98 & 86.60 & 53.52 & 68.27 \\
        LoRA-FISH-rk1-lay5 & 0.0142\% & 53.58 & 85.56 & 65.70 & 86.79 & 53.52 & 69.03 \\
        LoRA-FISH-rk1-lay6 & 0.0142\% & 53.06 & \textbf{87.02} & 65.70 & 86.96 & 53.52 & 69.25 \\
        LoRA-FISH-rk1-lay8 & 0.0142\% & 55.21 & 83.47 & 64.62 & 87.23 & 53.52 & 68.81 \\
        LoRA-FISH-rk1-lay10 & 0.0142\% & \textbf{56.32} & 85.74 & \textbf{66.43} & 87.21 & 53.52 & \textbf{69.84} \\
        LoRA-FISH-rk1-lay12 & 0.0142\% & 53.44 & 83.33 & 65.70 & \textbf{87.36} & 53.52 & 68.67 \\

        \hdashline
        Original-LoRA & 0.0184\% & 54.96 & 82.41 & 64.62 & 87.03 & 53.52 & 68.51 \\
        LoRA-FISH-rk1-lay5 & 0.0184\% & \textbf{55.96} & 84.18 & \textbf{67.51} & 86.81 & 53.52 & 69.60 \\
        LoRA-FISH-rk1-lay6 & 0.0184\% & 53.31 & 85.80 & 66.06 & 87.02 & 53.52 & 69.15 \\
        LoRA-FISH-rk1-lay8 & 0.0184\% & 53.67 & 85.63 & 65.70 & 87.25 & 53.52 & 69.15 \\
        LoRA-FISH-rk1-lay10 & 0.0184\% & 55.41 & \textbf{86.68} & 66.06 & 87.10 & 53.52 & 69.75 \\
        LoRA-FISH-rk1-lay12 & 0.0184\% & 55.91 & 86.30 & 66.06 & \textbf{87.34} & 53.52 & \textbf{69.83} \\

        \bottomrule[1.5pt]
    \end{tabular*}
    \caption{Performance of different PEFT methods on GLUE dataset. We set the LoRA rank to 1 and select different layers as trainable parameters.}
    \label{tab:loradifferentlayer}
\end{table*}

\begin{table*}[htbp]
    \small
    \begin{tabular*}{\textwidth}{p{4cm}cccccccccc}
        \toprule[1.5pt]
        \textbf{Method} & \textbf{Trainable Parameters} & \textbf{CoLA} & \textbf{MRPC} & \textbf{RTE} & \textbf{STS-B} & \textbf{WNLI} & \textbf{Avg} \\

        \midrule
        Original-LoRA & 0.0057\% & 43.27 & 80.94 & 58.48 & 84.94 & 53.52 & 64.23 \\
        LoRA-FISH-rk1-lay5 & 0.0057\% & 44.28 & 80.25 & 58.48 & 86.41 & 53.52 & 64.59 \\
        LoRA-FISH-rk2-lay5 & 0.0057\% & 44.98 & 83.46 & 62.45 & 86.18 & 53.52 & 66.12 \\
        LoRA-FISH-rk4-lay5 & 0.0057\% & 44.75 & 81.95 & 58.48 & 86.67 & 53.52 & 65.08 \\
        LoRA-FISH-rk8-lay5 & 0.0057\% & 44.13 & 81.28 & \textbf{66.06} & 86.94 & 53.52 & 66.39 \\
        LoRA-FISH-rk16-lay5 & 0.0057\% & 44.41 & 83.29 & 63.54 & 87.41 & 53.52 & 66.43 \\
        LoRA-FISH-rk32-lay5 & 0.0057\% & \textbf{48.09} & \textbf{85.31} & 65.34 & \textbf{87.50} & 53.52 & \textbf{67.95} \\

        \hdashline
        Original-LoRA & 0.0099\% & 48.02 & 82.17 & 62.82 & 86.06 & 53.52 & 66.52 \\
        LoRA-FISH-rk1-lay5 & 0.0099\% & 51.21 & \textbf{85.74} & \textbf{66.06} & 86.84 & 53.52 & \textbf{68.67} \\
        LoRA-FISH-rk2-lay5 & 0.0099\% & \textbf{51.56} & 82.61 & 62.45 & 86.72 & 53.52 & 67.37 \\
        LoRA-FISH-rk4-lay5 & 0.0099\% & 50.76 & 83.59 & 64.26 & 86.54 & 53.52 & 67.74 \\
        LoRA-FISH-rk8-lay5 & 0.0099\% & 49.95 & 83.79 & 65.70 & 87.04 & 53.52 & 68.00 \\
        LoRA-FISH-rk16-lay5 & 0.0099\% & 47.16 & 85.15 & 63.90 & 87.03 & 53.52 & 67.35 \\
        LoRA-FISH-rk32-lay5 & 0.0099\% & 51.01 & 85.24 & 64.98 & \textbf{87.61} & 53.52 & 68.47 \\

        \hdashline
        Original-LoRA & 0.0142\% & 51.87 & 84.35 & 64.98 & 86.60 & 53.52 & 68.27 \\
        LoRA-FISH-rk1-lay5 & 0.0142\% & \textbf{53.58} & 85.56 & 65.70 & 86.79 & 53.52 & \textbf{69.03} \\
        LoRA-FISH-rk2-lay5 & 0.0142\% & 52.56 & \textbf{85.70} & \textbf{66.06} & 86.50 & 53.52 & 68.87 \\
        LoRA-FISH-rk4-lay5 & 0.0142\% & 51.47 & 83.89 & 61.73 & 86.52 & 53.52 & 67.43 \\
        LoRA-FISH-rk8-lay5 & 0.0142\% & 50.00 & 85.42 & 64.98 & 86.93 & 53.52 & 68.17 \\
        LoRA-FISH-rk16-lay5 & 0.0142\% & 52.67 & 85.24 & 64.62 & 87.09 & 53.52 & 68.63 \\
        LoRA-FISH-rk32-lay5 & 0.0142\% & 52.15 & 84.26 & 64.98 & \textbf{87.42} & 53.52 & 68.47 \\

        \hdashline
        Original-LoRA & 0.0184\% & 54.96 & 82.41 & 64.62 & 87.03 & 53.52 & 68.51 \\
        LoRA-FISH-rk1-lay5 & 0.0184\% & \textbf{55.96} & 84.18 & 67.51 & 86.81 & 53.52 & \textbf{69.60} \\
        LoRA-FISH-rk2-lay5 & 0.0184\% & 53.21 & \textbf{85.10} & 68.95 & 86.65 & 53.52 & 69.49 \\
        LoRA-FISH-rk4-lay5 & 0.0184\% & 52.03 & 84.24 & 63.90 & 86.61 & 53.52 & 68.06 \\
        LoRA-FISH-rk8-lay5 & 0.0184\% & 50.67 & 84.70 & 64.98 & 86.87 & 53.52 & 68.15 \\
        LoRA-FISH-rk16-lay5 & 0.0184\% & 51.92 & 84.66 & \textbf{70.40} & 87.18 & 53.52 & 69.54 \\
        LoRA-FISH-rk32-lay5 & 0.0184\% & 53.17 & 83.91 & 67.15 & \textbf{87.47} & 53.52 & 69.04 \\

        \bottomrule[1.5pt]
    \end{tabular*}
    \caption{Performance of different PEFT methods on GLUE dataset. We set the trainable layers to 5 and select different LoRA ranks.}
    \label{tab:loradifferentrank}
\end{table*}

We select different trainable layers and LoRA ranks. The results in Tables \ref{tab:loradifferentlayer} and \ref{tab:loradifferentrank}.

\section{Hyperparameter Settings of the Experiment}
\label{sec:appendix3}

The hyperparameter settings in our experiment. The result in Table \ref{tab:hyperparameter}. 

\begin{table}[htbp]
  \begin{tabular}{l  l}
  \toprule[1.5pt]
  \textbf{Hyperparameters} & \textbf{value} \\
  \midrule

  \multirow{2}{*}{batch size} & 32 in classification task \\
   & 4 in generation task \\
  \midrule
    
  gradient accumu-  & 1 in classification task \\
  lation steps & 4 in generation task \\

  \midrule
  \multirow{3}{*}{learning rate} & 5e-5 in classification task  \\ 
  & 5e-6 in WNLI \\
  & 2e-5 in generation task \\

  \midrule
  \multirow{2}{*}{epoch} & 400 in classification task\\
   & 10 in generation task \\

  \midrule
  \multirow{2}{*}{early stop} & Yes in classification task\\
   & No in generation task \\

  \midrule
  optimizer & AdamW \\ 
  \midrule
  seed & 42 \\
  \midrule
  \multirow{2}{*}{warm up ratio} & 0.0 in classification task\\
   & 0.03 in generation task \\

  \midrule
  lr scheduler & linear in classification task \\
  type & cosine in generation task\\

  \midrule
  number of FISH & 128 in classification task \\
  Mask samples & 256 in generation task\\
  \midrule

  dataset used for & \multirow{2}{*}{ train set} \\
  Fisher estimation & \\
  \midrule

  Prefix-Tuning's & \multirow{2}{*}{ 30} \\
  prefix length $l$ & \\

  \midrule
  LoRA rank & details in Table \ref{tab:lorarank} \\
  \midrule
  LoRA $\alpha$ & 2 * rank \\
  \midrule
  LoRA dropout  & 0.1 in Qwen\\
  rate & 0.0 in other model \\

  \midrule


  Selected LoRA & \multirow{2}{*}{details in Table \ref{tab:selectedweights}} \\
  weights & \\

  \midrule
  distributed type & \multirow{2}{*}{zero-3 offload} \\
  (generation task) &  \\

  \midrule
  \multirow{2}{*}{Nvidia GPU} & 4 * A100 80G for Qwen \\
  & 8 * A5000 24G for others \\

  \midrule
  max sequence & \multirow{2}{*}{details in Table \ref{tab:detailGPT2}} \\
  length &  \\
  \bottomrule[1.5pt]
  \end{tabular}
  \caption{The hyperparameters in our experiment.}
  \label{tab:hyperparameter}
\end{table}

\begin{table}[htbp]
      \begin{tabular}{l l}
      \toprule[1.5pt]
        \textbf{Task-fixing the proportion} & \multirow{2}{*}{\textbf{LoRA Rank Value}} \\
        \textbf{of trainable parameters} & \\

        \midrule
        BERT-base-cased & 1 \\ 
        ModernBERT-base & 1 \\ 
        LLaMA-3.2-1B & 1 \\

        \midrule
        \textbf{Task-fixing the number} & \multirow{2}{*}{\textbf{LoRA Rank Value}} \\
        \textbf{of trainable layers} & \\

        \midrule
        BERT & 16  \\ 
        Phi-4-mini-instruct & 64  \\ 
        Qwen2.5-7B & 64  \\

        \bottomrule[1.5pt]
        \end{tabular}
        \caption{The different LoRA Rank value in different tasks.}
        \label{tab:lorarank}
\end{table}

\begin{table}[htbp]
      \begin{tabular}{l l}
      \toprule[1.5pt]
      \textbf{Pre-trained model} & \textbf{Selected LoRA Weights} \\
      \midrule
      BERT-base-cased & $W_Q$, $W_K$, $W_V$ \\ 
      ModernBERT-base & $W_O$ \\ 
      LLaMA-3.2-1B & $W_O$ \\ 
      Phi-4-mini-instruct & $W_Q$, $W_K$, $W_V$  \\ 
      Qwen2.5-7B & $W_Q$, $W_K$, $W_V$  \\

      \bottomrule[1.5pt]
      \end{tabular}
      \caption{The different selected LoRA weights in different pre-trained models.}
      \label{tab:selectedweights}
\end{table}

\begin{table}[htbp]
  \begin{tabular}{p{3cm} l}
  \toprule[1.5pt]
  \textbf{Task} & \textbf{max\_seq\_length} \\
  \midrule
  SST-2 & 128 \\ 
  CoLA & 128  \\ 
  RTE & 384 \\ 
  STS-B & 256 \\
  WNLI & 128 \\
  MRPC & 128 \\
  GSM8K & 512 \\
  \bottomrule[1.5pt]
  \end{tabular}
  \caption{The different max\_seq\_length in different tasks.}
  \label{tab:detailGPT2}
\end{table}

\section{Evaluation Metrics}
\label{sec:appendix33}

GLUE is a multi-task benchmark that contains 10 datasets for LLM evaluation. In our experiment, we select only the CoLA \cite{DBLP:journals/corr/abs-1805-12471}, 
MRPC \cite{DBLP:conf/acl-iwp/DolanB05}, RTE, SST-2 \cite{DBLP:conf/emnlp/SocherPWCMNP13}, STS-B \cite{DBLP:journals/corr/abs-1708-00055}, 
and WNLI \cite{DBLP:conf/kr/LevesqueDM12} datasets because the remaining datasets contain too much text and require excessive training time.  
Then we calculate the average score of these six datasets. Due to the limitation of uploading test set results to the official website only twice a day, we use only the validation set results. 

GSM8K is a dataset consisting of 8.5K high-quality grade school math problems designed to support multi-step reasoning tasks. 
It includes detailed step-by-step solutions for each problem, aiming to enhance the mathematical reasoning capabilities of machine learning models.

Different datasets have different evaluation metrics. The CoLA dataset uses Matthews correlation coefficient \cite{matthews1975comparison}. 
The STS-B uses Pearson and Spearman correlation coefficients. The MRPC uses a combined score (half the sum of $F_1$ and Accuracy). 
The rest of the datasets use Accuracy. 
We also analyze the average loss score of these six datasets.

\section{Pictures of Specific Results for Generation Task (Qwen2.5-7B)}
\label{sec:appendix4}

\begin{figure*}[htbp]
    \centering
    \centering
    \includegraphics[width=1.0\linewidth]{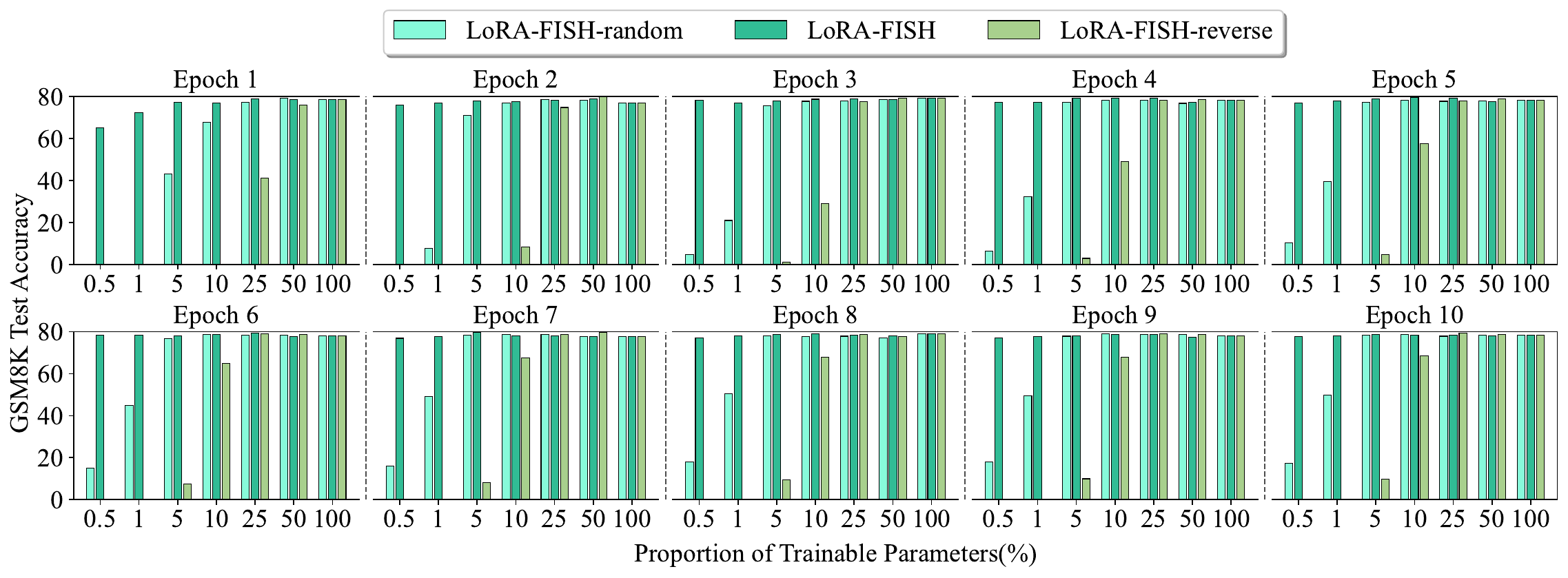}
    \caption{Performance of FISH-Tuning method in different proportion of Trainable Parameters on the GSM8K dataset using Qwen2.5-7B model
    for different training epochs.}
    \label{fig:epoqwenwhole}
\end{figure*}

\begin{figure*}[htbp]
    \centering
    \centering
    \includegraphics[width=1.0\linewidth]{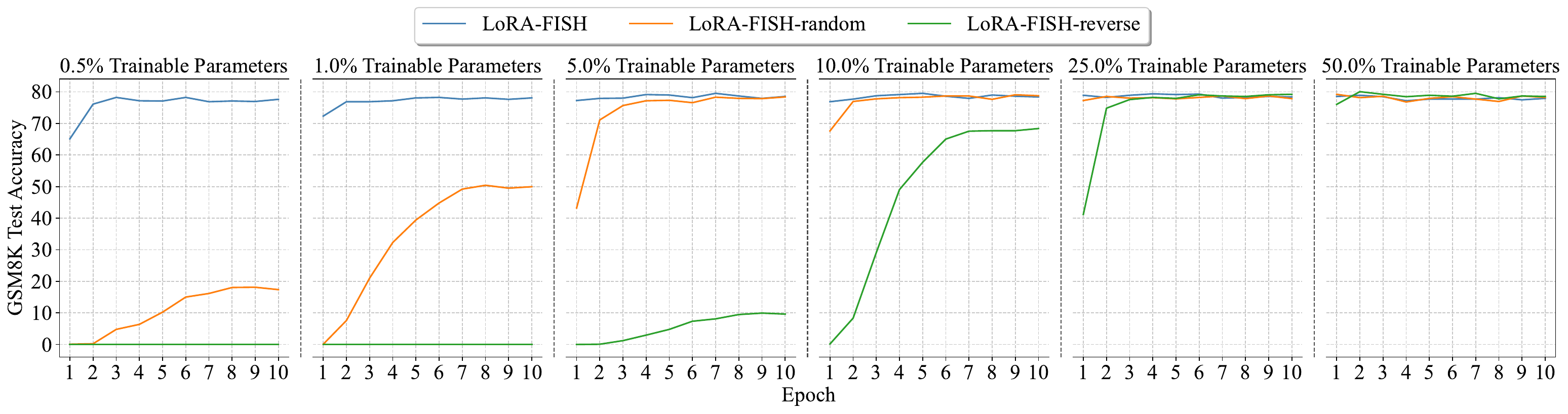}
    \caption{LoRA-FISH converges faster compared to LoRA-FISH-random and LoRA-FISH-reverse using Qwen2.5-7B model
    for different proportion of Trainable Parameters.}
    \label{fig:proqwenwhole}
\end{figure*}

Fig.~\ref{fig:epoqwenwhole} and Fig.~\ref{fig:proqwenwhole} show the results of different epochs and ratios in Qwen2.5-7B model.

\section{Pictures of Specific Results for Generation Task (Phi-4-mini-instruct)}
\label{sec:appendix5}

\begin{figure*}[htbp]
    \centering
    \centering
    \includegraphics[width=1.0\linewidth]{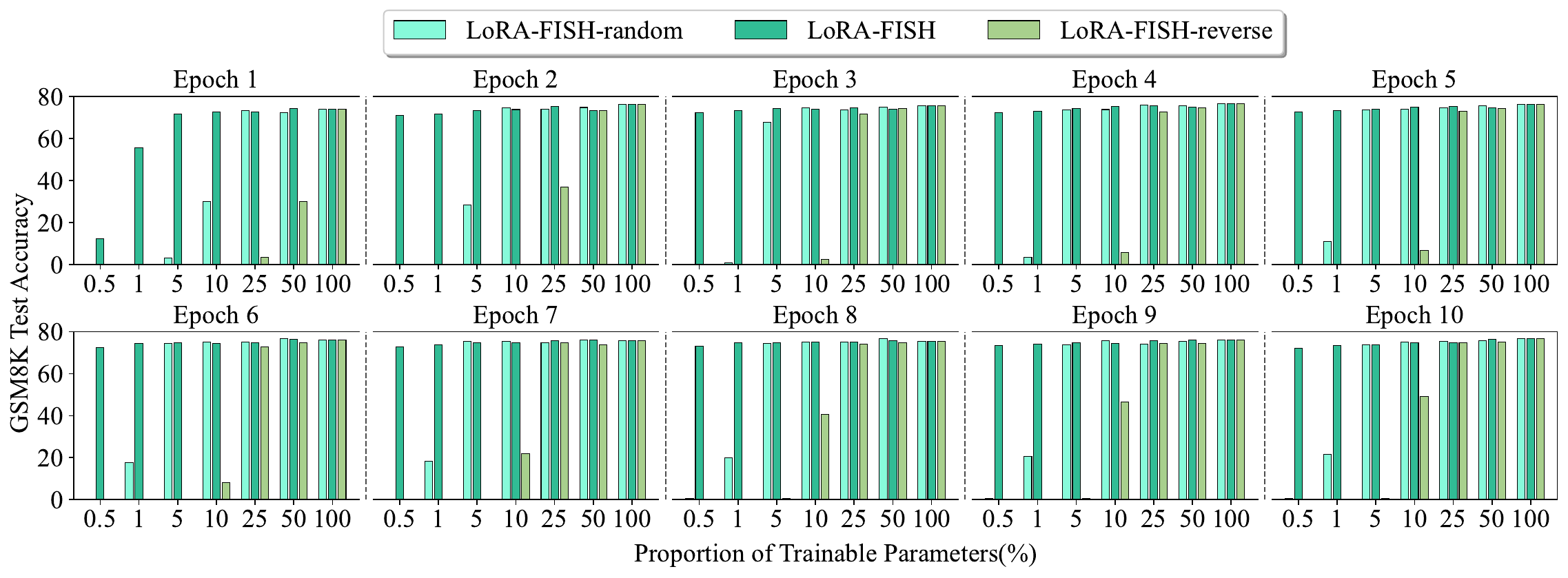}
    \caption{Performance of FISH-Tuning method in different proportion of Trainable Parameters on the GSM8K dataset using Phi-4-mini-instruct
    model for different training epochs.}
    \label{fig:epophiwhole}
\end{figure*}

\begin{figure*}[htbp]
    \centering
    \centering
    \includegraphics[width=1.0\linewidth]{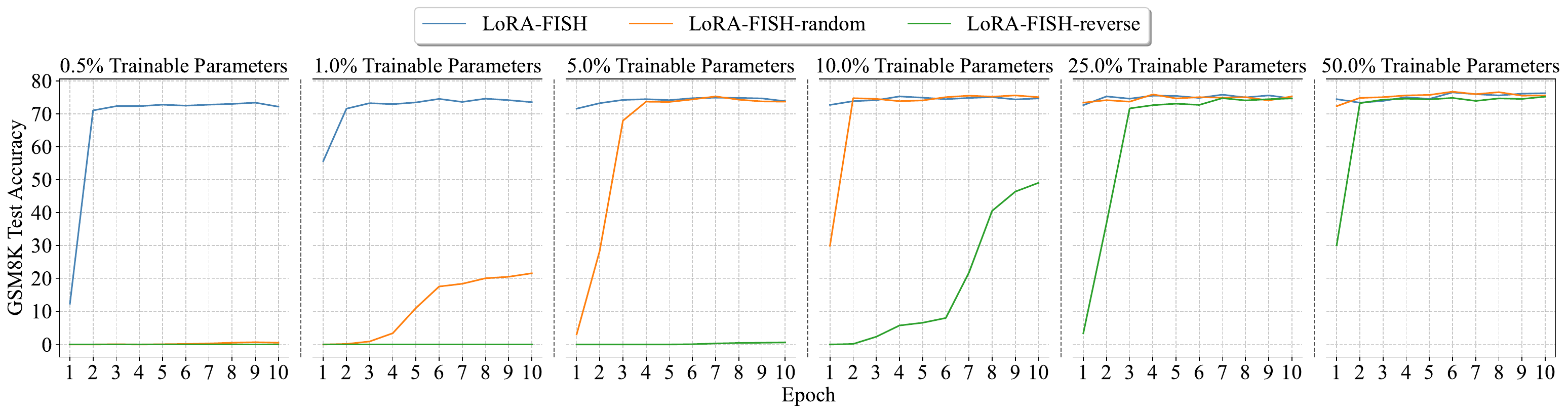}
    \caption{LoRA-FISH converges faster compared to LoRA-FISH-random and LoRA-FISH-reverse using Phi-4-mini-instruct model
    for different proportion of Trainable Parameters.}
    \label{fig:prophiwhole}
\end{figure*}

Fig.~\ref{fig:epophiwhole} and Fig.~\ref{fig:prophiwhole} show the results of different epochs and ratios in Phi-4-mini-instruct model.

\begin{figure*}[htbp]
    \centering
    \begin{minipage}[c]{1\linewidth}
        \centering
        \includegraphics[width=1.0\linewidth]{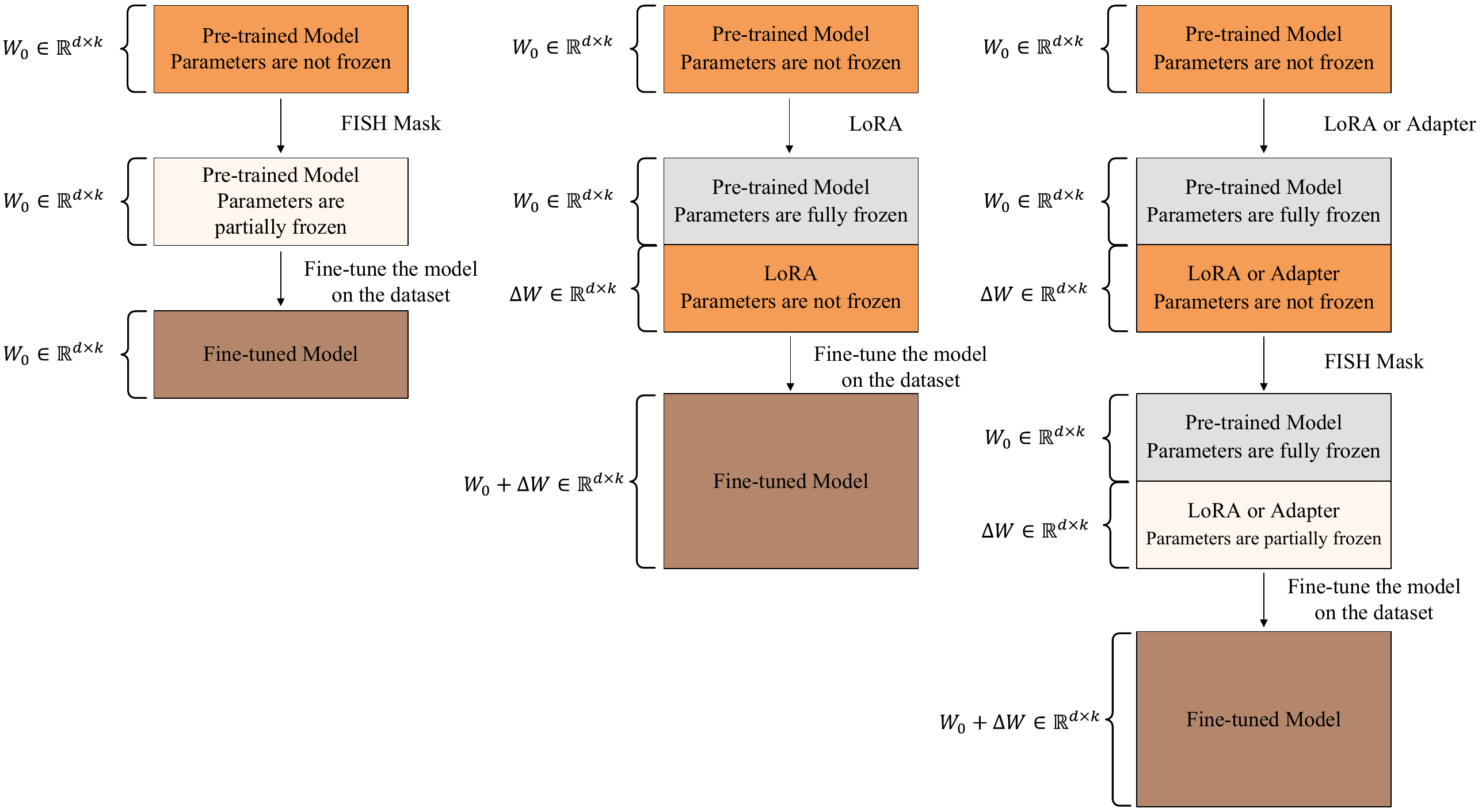}
    \end{minipage} 
    \caption{Original FISH Mask method (left) and original LoRA method (mid) and our FISH-Tuning method (right).}
    \label{fig:fishtuningintrowhole}
\end{figure*}

\end{document}